\documentclass[11pt]{article}

\usepackage{amsthm}
\usepackage{amssymb}
\usepackage{graphicx}
\usepackage{epstopdf}
\usepackage{amsmath,amstext,amssymb,amsfonts,verbatim}
\usepackage{xcolor}
\usepackage{hyperref}
\usepackage{algorithm,rotating}
\usepackage[noend]{algpseudocode}
\usepackage{geometry}
\hypersetup{colorlinks=false,linkbordercolor=red,linkcolor=blue,pdfborderstyle={/S/U/W 1}}

\theoremstyle{definition}
\newtheorem{defn}{Definition}[section]

\newcommand{\la}{\langle}
\newcommand{\ra}{\rangle}

\title{\vspace{-1cm} Algorithmic Causal Deconvolution of Intertwined Data and Networks by Generating Mechanism\thanks{An online implementation of the main ideas and techniques found in this paper can be accessed at \url{http://www.complexitycalculator.com/deconvolution}. Code in R and the Wolfram Language is freely available at \url{https://github.com/allgebrist/Causal-Deconvolution-of-Networks/}.}}

\author{Hector Zenil$^{1,2,3,5}$, Narsis A. Kiani$^{1,2,3}$, Allan A. Zea$^{1,4}$, Jesper Tegn\'er$^{2,5}$\\
$^1$ Algorithmic Dynamics Lab, Centre for Molecular Medicine,\\ Karolinska Institute, Stockholm, Sweden\\
$^2$ Unit of Computational Medicine, Department of Medicine,\\Karolinska Institute, Stockholm, Sweden \\
$^3$ Algorithmic Nature	Group, LABORES	for	the	Natural	and\\Digital Sciences, Paris, France\\
$^4$ Escuela de Matem\'atica, Facultad de Ciencias, UCV, Caracas, Venezuela\\
$^5$ Biological and Environmental Sciences and Engineering Division,\\	Computer, Electrical and Mathematical	Sciences and Engineering\\Division, King Abdullah University of Science and\\	Technology (KAUST), Kingdom of Saudi Arabia\\
{\{hector.zenil, narsis.kiani, jesper.tegner\}@ki.se}}

\date{}

\begin{document}

\maketitle

\begin{abstract}
Complex data usually results from 
the interaction of objects produced by 
different generating mechanisms. Here we introduce a universal, unsupervised and parameter-free model-oriented approach, based upon the seminal concept of \textit{algorithmic probability}, that decomposes an observation into its most likely algorithmic generative sources. Our approach uses a causal calculus to infer model representations. We demonstrate its ability to deconvolve interacting mechanisms regardless of whether the resultant objects are strings, space-time evolution diagrams, images or networks. While this is mostly a conceptual contribution and a novel framework, we provide numerical evidence evaluating the ability of our methods to separate data from observations produced by discrete dynamical systems such as cellular automata and complex networks. We think that these  separating techniques can contribute to tackling the challenge of causation, thus complementing  other statistically oriented approaches.\\
    
\end{abstract}

\newpage

\section{Introduction}

To extract representations leading to their generative mechanisms 
from data, especially without making arbitrary decisions based on biased assumptions, is a central challenge in most areas of scientific research, particularly 
given the major limitations of current machine and deep learning paradigms, which often lose sight of a model's components. 
Typically, models encode features of data in statistical form and in single variables, even in cases where several data sources are involved.

Broadly speaking, extracting candidate models and defining model-based approaches guided by data is one of the main challenges in the areas of machine learning, artificial intelligence and causal discovery. Here we introduce a framework based upon the theory of algorithmic probability, which in our formulation is capable of identifying different sources that may explain and provide different models for each of the possible sources of convoluted or intertwined data.

\subsection{Survey of Related Work}

Casual inference has been one of the most challenging problems in science. The debate about causality has not prevented the development of successful and mature mathematical and algorithmic frameworks, first in the form of logic and classical statistics, and today in the form of dynamical systems, computational mechanics, computability and algorithmic complexity. Based on mature mathematical notions that are acknowledged to fully characterise the concept of randomness, we introduced a suite of algorithms~\cite{maininfo} with which to study the algorithmic information dynamics of evolving systems, and methods to reduce the dimensions of data~\cite{mils} based on first principles. Algorithmic data dimension reduction and algorithmic deconvolution are challenges which can be viewed as opposite sides of the same coin. On the one hand, data reduction is achieved by finding elements that are considered redundant, using as a criterion their contribution to the algorithmic content of the description of the data. On the other hand, deconvolution by generative source is perhaps the ultimate goal of partition, clustering and machine learning algorithms. However, these approaches often lose sight of their goal of causal decomposition and rather seek to identify common features of data as evidence of the possible common origin of said data. 
For example, in signal processing, popular methods such as $k$-means~\cite{lloyd} or $k$-medoids~\cite{Kaufman} define heuristics based on the minimisation of distance among data points according to some metric. Some other popular methods, such as support vector clustering~\cite{siegelmann} and traditional machine learning techniques~\cite{kubat}, draw on probability distributions, regression, and correlation techniques, providing means for linear separation producing different groups. This includes deep neural networks~\cite{torr} based on constructing a differentiable landscape on which elements are statistically mapped for classification purposes. Another type of separating method applied to objects such as graphs relies on graph-theoretic properties (e.g. ~\cite{villani}) shared among networks. In this category belong ways to separate graphs by indices such as edge betweenness or by  the frequency of over-representation of certain subgraphs, also called network motifs~\cite{alon,alon2}, and by more sophisticated criteria such as shared graph spectral features~\cite{newman,bencz,spielman,spielman2}. All these methods make the assumption that, by virtue of objects sharing statistical, topological or algebraic features~\cite{liu}, the said objects may be generated by the same means or from the same sources. 

Classical information theory has provided  ways to capture and encode statistical properties from data, and these have made an impact on many areas of science. For example, mutual information can capture various averages based on associated distributions of statistical properties that are contained in one variable about another, that is, how information can be combined and decomposed in purely statistical terms. Recently, some methods based on information decomposition have been introduced~\cite{williams,ince} with the purpose of separating multivariate signals into their alleged generative sources. A recent proposal that has gained some traction is the so-called Partial Information Decomposition or PID~\cite{williams}, that falls short~\cite{ince}, among other reasons, because it can only tell what a variable can statistically tell about some other variable.

Thus we believe that there is a strong need to advance methods to decompose intertwined data coming from one or more sources that go beyond traditional statistics. Such methods would require novel approaches with more powerful indices---which also makes them more difficult to calculate. We call our method \textit{causal deconvolution by generative source} and offer it as an alternative to statistical inference approaches such as PID, which are hamstrung by limitations. This is particularly important in applications to areas such as biology or neuroscience, where the ability of a system to decompose multiple information sources in a non-trivial fashion represents evolutionary and cognitive advantages independent of classical information-theoretic constraints. Quantitative measures to disentangle complex sources and new methods to help tell apart fused mechanisms from other signals or simply to tell noise from signals are thus of broad interest in areas where causal analysis is germane.

Notably related to the kind of deconvolution and decomposition explored in this paper are methods based on pattern recognition~\cite{watanabe}, classical information theory~\cite{williams} (a survey can be found in~\cite{lizier}) and lossless compression\cite{liA}, the most relevant methods being the ones based on information distances~\cite{li} and compression, such as the so-called Normalised Information Distance~\cite{bennett}, and its related measure the Normalised Compression Distance~\cite{li2}, and other variations. All these methods can also be adapted to make use of other complexity indices or lossless compression algorithms.

One criticism levelled at several, if not most, algorithms and complexity measures is that in their estimation of some complexity index, the methods only assign a number to data from which nothing else can be extracted and has very little value. An example would be a compressed file,
because a compressed file is not only a black box which is almost impossible to decipher, but it is also not a model, rule or computer program that can receive inputs or be run for a larger number of steps to produce more data or make predictions. Along the same lines is Rissanen's~\cite{rissanen} Minimum Description Length, inspired by Solomonoff's induction method of algorithmic probability~\cite{solomonoff}, 
which eschews the strength of a Turing-complete language and elects to compress data using weaker models of computation.

In attempting to address the drawbacks of statistical inference, methods such as inductive inference~\cite{gulwani}, inductive programming, and program synthesis have been advanced. In this direction, the concept of algorithmic probability introduced by Solomonoff~\cite{solomonoff} was proven to be optimal and universal, but its semi-computability was a deterrent to its application, leading some researchers, led by Crutchfield et al. to circumvent it and instead use computationally constrained models such as computational mechanics~\cite{crutch1,crutch2}. 

In the approach followed in this paper, which can be thought of as related to computational mechanics, we replace the methods used in computational mechanics based on, e.g., Markov processes and Bayesian inference, by a measure based on algorithmic probability and an empirical estimation of the Universal Distribution (the distribution associated with algorithmic probability), while staying within the boundaries of
the field of computational mechanics itself. In a similar category of inductive inference, but mostly of a theoretical nature is AIXI as introduced by Hutter~\cite{hutter}, combining algorithmic probability with decision theory by way of 
Bayes' theorem, replacing the prior with the Universal Distribution as a prior. It is similar to Levin's search~\cite{levin2}, that is,it dovetails Turing machines (interleaves computer programs one step at a time from shortest to longest). It is designed particularly for applications to reinforcement learning. In current deployments AIXI circumvents uncomputability and intractability by relying on popular compression algorithms such as LZW, Minimum Description Length, Monte Carlo search and Markov processes, thereby effectively using weaker models of computation. In another category are methods introduced by Hern\'andez-Orallo et al. and their computational measures of information gain and reinforcement in inference processes~\cite{hernandez2000,hernandez1999}, alternatives to ours. One novelty in our approach based on the concept of algorithmic information dynamics~\cite{maininfo,zenilevo} is the precomputation of a very large set of small models able to explain small pieces of data, which assembled together in sequence, can build a full model of larger data. The method's precomputation allows practical applications in linear time by implementing a look-up table~\cite{d4,d5}, which combined with classical information theory provides key hints on the algorithmically random versus non-random nature of data. It has proven able to deal with features that are not only statistical in nature, features captured by other methods such as Shannon entropy, pattern recognition, or lossless compression, but also with more convoluted features of an algorithmic nature that weaker computational approaches would miss~\cite{emergence}. By convoluted, convolution and deconvolution, we mean the original meaning of these words and not necessarily to the current field of convolutional neural networks. However, our approach can help understand and even help current deep learning techniques, including areas of convolution and deconvolution. A convolutional neural network (CNN) combines a set of primitive features extracted from data for classification purposes, deconvolving would involve opening the CNN and separating features by their most likely common generative sources.

Our approach builds upon previous work but is also based on our own work combined with the seminal ideas on counterfactuals of Judea Pearl et al.  and with their interventionist do-calculus~\cite{pearl}. Pearl's et al. interventionist calculus is a part of his theory of probabilistic causality, itself part of the study of Bayesian networks. Our approach is a complete bottom-up approach based on algorithmic probability~\cite{solomonoff,chaitin,levin}, similar in nature to AIXI and alternative to or within the boundaries of computational mechanics but conceived to be practical from the start and designed for immediate application, without compromising on the power of the computational model used for the inductive inference. The deconvolution method introduced here is based on our own algorithmic causal calculus~\cite{maininfo} or algorithmic information dynamics and involves finding the most likely (and thus shortest) generating mechanisms (computer programs) capable of reproducing an observation (data). For some other examples of areas in which these methods have found applications and been demonstrated to outperform computable measures see Refs.~\cite{emergence}, ~\cite{bdm}, and~\cite{numerical}, and for a non-trivial example in which entropic measures fail (by offering divergent descriptions of the same evolving system) see Ref.~\cite{zkgraph}. Moreover, behind the number or sequence of numbers matching observation/data and complexity, we also offer access to the rules generating the data that represent the generative model of the said data, that can thus be used for validation against present and future data, allowing predictions.

\section{Methods and Algorithms}

Cellular automata offer an optimal testbed because they are discrete dynamical systems able to illustrate an algorithm's inner workings because of their visual nature. They can be interpreted as 1-dimensional objects that produce 2-dimensional images when adding runtime, producing highly integrated 2-dimensional objects whose rows are strongly causally connected and are thus ideal testing cases. This does not mean, however, that the same methods cannot be applied to other images. After CA evolutions We then move to more applications demonstrating the method's capabilities on other objects of convoluted nature, such as complex networks.

The main intuition behind our algorithms is as follows. We look for pieces of observed data that may come from the same source or generating mechanism using as a guide the length of the set of possible computer programs that produce different pieces of the data when decomposed. The main point is that if a computer program generates the data, different regions of the data would be explained by the same algorithm and that algorithm will also have the same program length for regions coming from the same generating mechanism.

\subsection{Cellular Automata}

A cellular automaton is a computer program that applies in parallel a global rule composed of local rules on a tape of cells with symbols (e.g. binary). Thoroughly studied in~\cite{nks}, Elementary Cellular Automata (or ECA) are one-dimensional cellular automata that take into consideration in their local rules the cell next to the centre and the centre cell.

\begin{defn} A {\it cellular automaton} (or CA) is a tuple $\langle S, (\mathbb{L}, +), T, f \rangle$ 
with a set $S$ of states, a lattice $\mathbb{L}$ with a binary operation $+$, a neighbourhood template $T$, and a local rule $f$.
\end{defn}

The {\it set of states} $S$ is a finite set with elements $s$ taken from a finite alphabet $\sum$ with at least 2 elements. 

\begin{defn}
The {\it neighbourhood template} $T=\langle\eta_1, \dots,\eta_m\rangle$ is a sequence of $\mathbb{L}$. In particular, the neighbourhood of cell $i$ is given by adding the cell $i$ to each element of the template $T$: $T=\langle i+\eta_1, \ldots,i+\eta_m\rangle$. 
Each cell $i$ of the CA is in a particular state $c[i] \in S$. A {\it configuration} of the CA is a function $c: \mathbb{L} \rightarrow S$. The {\it set of all possible configurations} of the CA is defined as $S_\mathbb{L}$.
\end{defn}

As a discrete dynamical system, the {\it evolution of the CA} occurs in discrete time steps $t=0,1,2, \ldots,n$. The transition from a configuration $c_t$ at time $t$ to the configuration $c_{(t+1)}$ at time $t+1$ is induced by applying the local rule $f$. The local rule is to be taken as a function $f: S^{|T|} \rightarrow S$ which maps the states of the neighbourhood cells of time step $t$ in the neighbourhood template $T$ to cell states of the configuration at time step $t+1$:

\begin{equation}
c_{t+1}[i]=f\left(c_t[i+\eta_1], \dots,c_t[i+\eta_m]\right )
\end{equation}
The general transition from configuration to configuration is called the {\it global map} and is defined as: $F: S^\mathbb{L} \rightarrow S^\mathbb{L}$. The code in Wolfram Language is available in the Sup. Inf.

\subsection{Enumeration of ECA rules}

In the case of 1-dimensional CA, it is common to introduce the {\it radius} of the neighbourhood template which can be written as $\langle -r,-r+1, \ldots,r-1,r \rangle$ and has length $2 r+1$ cells. With a given radius $r$ the local rule is a function $f: \mathbb{Z}_{|S|}^{{|S|}^{(2r+1)}} \rightarrow \mathbb{Z}_{|S|}$ with $\mathbb{Z}_{|S|}^{{|S|}^{(2r+1)}}$ rules. 
\textit{Elementary Cellular Automata} or ECA, have a radius $r=1$ (closest neighbours), having the neighbourhood template $\langle -1,0,1\rangle$, meaning that the neighbourhood comprises a central cell. From this it follows that the rule space for ECA contains $2^{2^{3}}=256$ rules.

It is common to follow a lexicographic ordering scheme in the enumeration of CA and ECA as introduced by Wolfram~\cite{nks}. According to this scheme, the 256 ECA rules can be encoded by only 8-bits once part of the rule is fixed for all of them.

\subsection{Randomly Interacting Cellular Automata}

The use of interacting programs such as cellular automata as examples illustrating our algorithms requires us to define how the interaction happens. That is, how it is decided what set of rules apply at the intersection of the interacting CA. For instance, one of the 2 sets of local rules or a 3rd set of rules effectively defines a super cellular automaton that most likely is another cellular automaton in a larger rule-space (requiring more states to define the 2 sub-cellular automata and the interaction).

Our interacting Cellular Automaton model, as introduced in~\cite{adams}, is of such a nature that only one of then survives the contact between the CA colours, or else they both disappear. This means that there are only 3 possible solutions to what happens when cells of both CA come into contact in the same neighbourhood, and that there is in reality a single controlling CA with 3 colours that governs the interaction of the other 2 whose local rules dictate that: grey survives, black survives or there only remains a white cell. However, grey does not need to be displayed because it is only auxiliary and determines what happens when cells of the different CA come in contact in the same neighbourhood. 

In particular we have it that $c_{t+1}(x_n)$ should be either white or black in case none of $c_{t}(x_{n-1})$, $c_{t}(x_{n})$ and $c_{t}(x_{n+1})$ is grey. Likewise, $c_{t+1}(x_n)$ should be either white or grey in case none of $c_{t}(x_{n-1})$, $c_{t}(x_{n})$ and $c_{t}(x_{n+1})$ is black. For example, let $N=\la c_t(x_{j-1}), c_t(x_j) ,c_t(x_{j+1})\ra$ and $M=\la c_{t'}(x_{i-1}), c_{t'}(x_i) , c_{t'}(x_{i+1}) \ra$ be 2 different mixed neighbourhoods. Then there is no correlation between the random values of $c_{t+1}(x_j)$ and of $c_{t'+1}(x_i)$. Note that we impose this independence both in case $N=M$ (so that the difference is only reflected in either the location (that is $i\neq j$) or in the time ($t\neq t'$)) and in case $N\neq M$. In particular the mixed neighbourhood $\la 2,2,1 \ra$ may sometimes yield a $0$, sometimes a $1$ and at yet other times a $2$.

In~\cite{JoostenGNS} and~\cite{JoostenFittest} these interactions are investigated in terms of evolution and complexity. In the Sup. Inf. we also provide 2 more examples of interacting CA and a comparison with other methods and measures, in particular the recently introduced Partial Information Decomposition~\cite{williams} using both classical Mutual Information~\cite{williams} and the Normalized Compression Distance~\cite{li2}. One striking difference with these other methods, is that ours does not require any associated probability distributions or arbitrary parameter choice, including the terminating criterion explained in Subsection~\ref{terminating}.

\subsection{Graph Complexity}\label{graphcomp}

The concept of \textit{Algorithmic Probability} (and of Levin's semi-measure, and Universal Distribution associated with it) has been introduced as a method for approximating algorithmic complexity based on the frequency of the patterns occurring in the adjacency matrix of a network. The measure applied to labelled graphs has been proven to be a tight upper bound of the algorithmic complexity of unlabelled graphs and therefore quite invariant to particular adjacency matrix choice~\cite{zenilkianitegner}.

More precisely, the algorithmic probability~\cite{solomonoff,levin,chaitin} of a subgraph $H \subseteq G$ is a measure of algorithmic probability based on the frequency of a random computer program $p$ producing $H$ when run on a 2-dimensional tape universal (prefix-free\footnote{The group of valid programs forms a prefix-free set (no element is a prefix of any other, a property necessary to keep $0 < m(G) < 1$). Because $m(G) < 1$, $m(s)$ is called a semi-measure, because not all programs halt and thus it never reaches 1.}) Turing machine $U$ also referred to as a \textit{Turmite}~\cite{langton}. That is, 
$$m(G) = \sum_{p:U(p) = H \subseteq G} 1/2^{|p|}$$.

The probability semi-measure $m(G)$ is related to algorithmic complexity $C(G)$ in that $m(G)$ is at least the maximum term in the summation of programs $m(G)\geq2^{-C(G)}$, given that the shortest program carries the greatest weight in the sum. 

The algorithmic complexity (also known as Kolmogorov-Chaitin complexity)~\cite{kolmo,chaitin} is the length of the shortest computer program that reproduces the data from its compressed form when running on a universal Turing machine.

The Coding Theorem~\cite{solomonoff,levin} establishes the connection between $m(G)$ and $C(G)$ as $|-\log_2 m(G) - C(G)| < c$, where $c$ is some fixed constant independent of $s$. The theorem implies that one can estimate the algorithmic complexity of a graph from the frequency of production by running random programs and applying the Coding theorem: $C(G)=-\log_2 m(G) + c$. We call this approach the Coding Theorem Method (CTM). The Coding theorem establishes that graphs produced with lower frequency by random computer programs have higher algorithmic complexity, and vice versa. Applying the so-called Coding Theorem Method (CTM) and Block Decomposition Method (BDM), as introduced in~\cite{d4,d5,kolmo2d,bdm}, based on estimations to algorithmic probability involves running a very large number of small computer programs according to a quasi-lexicographic order (from smaller to larger program size) to produce an empirical approximation to the Universal Distribution $m$. The BDM of a graph then consists in decomposing the adjacency matrix of a graph into subgraphs of sizes for which complexity values estimated by CTM are available, then reconstructing a sequence model that can explain the full $G$ by assembling in sequence the smaller models that produce all the parts of $G$ in combination with rules from classical information theory, as follows:

\begin{equation}
\label{newecaeq}
C(G) = \sum_{(r_u,n_u)\in Adj(G)_{d\times d}} \log_2(n_u)+C(r_u)
\end{equation}

\noindent where $Adj(G)_{d\times d}$ represents the set with elements $(r_u,n_u)$, obtained when decomposing the adjacency matrix of $G$ into all subgraphs of size $d$ contained in $G$. In each $(r_u,n_u)$ pair, $r_u$ is one such submatrix of the adjacency matrix and $n_u$ its multiplicity (number of occurrences). As can be seen from the formula, repeated subgraphs only contribute to the complexity value with the subgraph BDM complexity value once plus a logarithmic term as a function of the number of occurrences. This is because the information content of subgraphs is only sub-additive, as one would expect from the growth of their description lengths. Applications of $m(G)$ and $C(G)$ have been explored in~\cite{d4,d5,kolmo2d}, and include applications to graph theory and complex networks~\cite{zenilgraph} and in~\cite{kolmo2d} where the technique was first introduced.

The only parameter used for the application of BDM, as suggested in~\cite{bdm}, is to set the overlapping of the decomposition to the maximum 12 bits for strings and 4 square bits for arrays, given the current best CTM approximations~\cite{d5} from an empirical distribution based on all Turing machines with up to 5 states, with no string/array overlapping in the decomposition for maximum efficiency (as it runs in linear time), and for which the error (due to boundary conditions) has been shown to be bounded~\cite{bdm}.

However, the algorithm introduced here is independent of the method used to approximate algorithmic complexity, such as BDM. BDM assigns an index associated with the size of the most likely generating mechanism producing the data according to Algorithmic Probability~\cite{solomonoff}. BDM is capable of capturing features in data beyond statistical properties~\cite{bdm,zkgraph}, and thus represents an improvement over classical information theory. Because finding the program that reproduces a large object is computationally very expensive---even to approximate---BDM finds short candidate programs using another method~\cite{d4,d5} that finds and reproduces fragments of the original object and then puts them together as a candidate algorithmic model of the whole object~\cite{bdm,kolmo2d}. These short computer programs are effectively candidate models explaining each fragment, with the long finite sequence of short models being itself a generating mechanistic model. 

\subsection{Deconvolution Algorithms}

The aim of the deconvolution algorithm is to break a dataset into groups that do not share certain features (essentially  causal clustering and algorithmic partition by probable generative mechanism, completely different from traditional clustering and partition in machine learning approaches). Usually these characteristics are a parameter to maximise, but ultimately the purpose is to distinguish components that are generated similarly from those that are generated differently. In information-theoretic terms the question is therefore as follows: What are the elements (e.g. nodes or edges) that can break a network into the components that maximise their algorithmic information content, that is, those elements that preserve the information about the underlying programs generating the data?

Let $G$ be a graph and let $E=E(G)$ denote its set of edges. Let $G\backslash e$ denote the graph obtained after deleting an edge $e$ from $G$. The {\it information contribution} of $e$ to $G$ is given by $I(G,e):=C(G)-C(G\backslash e)$. A positive information contribution corresponds to information loss and a negative contribution to information gain. Here we wish to find the subset $F\subseteq E$ such that the removal of the edges in $F$ disconnects $G$ into $N$ components and minimises the loss of information among all subsets of edges, i.e. the subset such that $I(G,F)\leq I(G,S)$ for all $S\subseteq E$. Let us denote the number of connected components of $G$ by $k(G)$. Algorithm \ref{deconv} allows us to obtain the subgraph $(V,E\backslash F)$ subject to the above conditions. The desired subset of edges is then given by $F=E(G)\backslash E(\textnormal{\textsc{Deconvolve}}(G,N))$.

\begin{algorithm}[ht!]
\caption{Causal deconvolution algorithm}\label{deconv}
\begin{algorithmic}[1]
\Function{Deconvolve}{$G,N$}, $1\leq k(G)\leq N\leq |V(G)|$
\While{$k(G)<N$}
\State $\textit{informationLoss} \gets \varnothing$
\Statex\hspace{3em}{\color{gray} // for each edge $e$}
\For{$e\in E(G)$}
\If{$I(G,e)>0$}
\Statex\hspace{6em}{\color{gray} // store information contribution into {\it informationLoss}}
\State $\textit{informationLoss} \gets \textit{informationLoss} \cup \{I(G,e)\}$
\EndIf
\EndFor
\Statex\hspace{3em}{\color{gray} // calculate the minimal information loss across all edges}
\State $\textit{minLoss} \gets \min({informationLoss})$
\Statex\hspace{3em}{\color{gray} // remove all candidate edges from $G$}
\State $\textit{G} \gets G\backslash\{e\in E(G):I(G,e)=\textit{minLoss}\}$
\EndWhile
\Return{$G$}
\EndFunction
\end{algorithmic}
\end{algorithm}

The only parameter that Algorithm~\ref{deconv} requires is the number of components into which an object will be decomposed. However, there is a natural way to find the optimal terminating step and therefore the number of maximum possible components that minimise the sum of the lengths of the candidate generating mechanisms, making the algorithm truly parameter-free, as it is not required to have a preset number of desired components. $N$ should be chosen to be equal to the maximum number of components into which the graph should be broken. However, an alternative Algorithm~\ref{deconvterm}, determines the optimal number of components and requires no $N$ parameter.

Before introducing the terminating criterion (c.f. next section) for the number of components, let's analyse what it might mean for 2 components $s_1$ and $s_2$ to have the same algorithmic information content $C(s_1)=C(s_2)$. Clearly that subcomponents $s_1$ and $s_2$ have the same algorithmic complexity (an integer---or a real value if using AP-based BDM---indicating the size of the approximated minimal program) does not imply that the 2 components are generated by exactly the same generating mechanism. However, because of the exponential decay of the algorithmic probability of an increasingly random object, we have it that the less random it is, the exponentially more likely it is that the underlying mechanism will be the same (see Fig.~\ref{x}C). This is because there are exponentially fewer short programs than long ones. For example, in the extreme case of connected graphs, we have it that the complete graph denoted by $K_n$ has the smallest possible algorithmic complexity $\sim \log(n)$. If $C(s_1) = C(s_2) \sim \log(n)$ then $s_1$ and $s_2$ are, with extremely high probability, generated by the same algorithm that generates either the complete graph or the empty graph (with the same lowest algorithmic complexity, as it requires no description other than either all nodes connected or all nodes disconnected). Conversely, if $C(s_1)=C(s_2)$ but $C(s_2)$ and $C(s_1)$ depart from $\log(n)$ (and approximate algorithmic randomness), then the likelihood of being generated by the same algorithm exponentially vanishes. So the information regarding both the algorithmic complexity of the components and their relative size sheds light on the candidate generating mechanisms and is less likely to coincide `by chance' for non-trivial cases.

\subsubsection{Algorithm Terminating Criterion}
\label{terminating}

The immediate question is where we should stop breaking down a system into its causal components. The previous section suggests a terminating criterion. Let $S$ be the object which has been produced by $N$ mostly independent generative mechanisms. We decompose $S$ into $n$ parts $s_1, \ldots, s_n$ in such a way that each $s_i$, $i\in\{ 1\ldots n\}$ has an underlying generating mechanism found by running the algorithm iteratively for increasing $n$, But after each iteration we calculate the minimum of the differences in algorithmic complexity among all subcomponents. The algorithm should then stop where the number of subcomponents is exactly $N$ when the sum of the lengths---the estimated algorithmic complexity---of each of the programs will diverge from the expected $log(N)$ because the length of the individual causal mechanisms producing each new component will be breaking a component that could previously be explained by the causal mechanism at a previous iteration of the algorithm. An implementation of this idea for a graph is shown in Algorithm~\ref{deconvterm}.

As a trivial example, let's take the string $1^n$, where $S^n$ means that the pattern $S$ is repeated $n$ times. After application of the algorithm, the terminating criterion will suggest that $1^n$ cannot be broken down into smaller segments, each with a different causal generating mechanism, the sum of whose total length will be shorter than the length of the generating mechanism producing $1^n$ itself. This is because the sum of the length of the shortest programs $\sum_i |p_i|$ running on a universal Turing machine generating segments of $1^n$ of length $m_i<n$ each, such that the concatenation $\cup_{i=1} p_i= 1^n$, will be strictly greater than $C(1^n)$, given that each $p_i$ halting criterion will require $i \log m_i$ bits more than $C(1^n)$.

In the case of Fig.~\ref{b}, the terminating criterion retrieves $N=3$ components from the 2 interacting ECA (rule 60 and 110). This does not contradict the fact that we started from 2 generating mechanisms, because there are 3 clear regimes that are actually likely to be reproducible by 3 different generating mechanisms, as suggested by the deconvolution algorithm itself, and as found in~\cite{jurgenzenil}, where it has been shown that rule 110 can be emulated by the composition of 2 simpler ECA rules (rules 51 and 118). As seen in Fig.~\ref{b}, among the possible causal partitions, $N=2$ successfully deconvolves ECA rule 60 from rule 110 on the first run, with a stronger difference than the difference found between $N=3$ components when breaking rule 110 into its 2 different regimes.

\begin{algorithm}[ht!]
\caption{Causal deconvolution with terminating criterion}\label{deconvterm}
\begin{algorithmic}[1]
\Function{Deconvolve}{$G,\varepsilon$},
    \State $\textit{informationLoss} \gets \varnothing$
    \Statex\hspace{1.5em}{\color{gray} // for each edge $e$}
    \For{$e\in E(G)$}
        \If{$I(G,e)>0$}
            \Statex\hspace{4.5em}{\color{gray} // store information contribution into {\it informationLoss}}
            \State $\textit{informationLoss} \gets \textit{informationLoss} \cup \{I(G,e)\}$
        \EndIf
    \EndFor
    \For{$\textit{loss}\in\textit{informationLoss}$}
        \State $difference \gets 0$
        \If{$|\textit{informationLoss}|>1$}
            \Statex\hspace{4.5em}{\color{gray} // copy maximum loss to {\it maxLoss}}
            \State $\textit{maxLoss} \gets \max(\textit{informationLoss})$
            \State $\textit{informationLoss} \gets \textit{informationLoss}\backslash\{\textit{maxLoss}\}$
            \Statex\hspace{4.5em}{\color{gray} // calculate difference between the old and new maxima}
            \State $\textit{difference} \gets \textit{maxLoss}-\max(\textit{informationLoss})$
            \Statex\hspace{4.5em}{\color{gray} // if {\it difference} significantly departs from $\log(2)$}
            \If{$|\textit{difference}-\log(2)|>\varepsilon$}
                \Statex\hspace{6em}{\color{gray} // remove all candidate edges from $G$}
                \State $G \gets G\backslash\{e\in E(G):I(G,e)=\max(\textit{informationLoss})\}$
            \EndIf
        \EndIf
    \EndFor
    \Return{$G$}
\EndFunction
\end{algorithmic}
\end{algorithm}

The term $\varepsilon$ is related to the number of components $N$ from Algorithm~\ref{deconv}, or rather substitutes for $N$. $\varepsilon$ is an auxiliary cutoff value that determines when to stop the algorithm without making an arbitrary choice of number of subcomponents $N$. $\varepsilon$ is 0 for the theoretical cutoff value $\log(2)$ that determines the effect of perturbations performed on the network. If removing an edge has an effect greater than $\log(2)$, then such an edge does not belong to the same underlying algorithm explaining the rest of $G$, given that the program size of a deterministic object generated by the same computer program does not change by more than $\log(2)$ and thus nor does its algorithmic complexity. In contrast, if the perturbation by edge removal has a loss of at most $\log(2)$ bits, then it means that it is likely to be reconstructed by the original computer program because $\log(2)$ is the growth in the description of a computer program (or a deterministic system) that accounts only for running time. In other words, if the perturbation is above $\log(2)$, it means that such an edge may have disconnected 2 or more causally independent components with different computer programs likely able to explain each different subcomponent with fewer bits than when keeping those components together using such an edge. 

In practice, however, such a strict cutoff value does not occur, so $\varepsilon$ accounts for a difference or error not far from $\log(2)$. Moreover, $\varepsilon$ can be estimated from the sequential information differences calculated from the absolute distances between the differences of consecutive values in the information signature (the list of information values for all edges sorted by maximum contribution) and its deviation from $\log(2)$, so no cut is made for an edge with information difference below $\log(2) + \varepsilon$  (see Fig.~\ref{c2}).

\subsubsection{Time Complexity}

The algorithms for network deconvolution introduced in this section run in polynomial time. Let $M$ denote the number of edges of the graph $G$. The brute force algorithm for this problem searches the edge such that its removal minimises the loss of information and deletes it, repeating this process for all edges of $G$ until $N$ subcomponents are reached, which has a worst-case time complexity of $O(M^2)$. Algorithm~\ref{deconv} is different from the brute force approach in that edges with equal minimal contributions to the loss of information are not deleted sequentially but all at once, but its time complexity is also of $O(M^2)$. The outer loop that verifies whether the number of desired subcomponents is reached is removed in Algorithm~\ref{deconvterm}, allowing us to find the optimal terminating step of the deconvolution in time $O(M)$.

\section{Numerical Experiments}

Behind our deconvolution methods is the idea that we can find a set of small computer rules or programs able to reconstruct a piece of data, Figs.~\ref{a}C-D illustrate this. In the deconvolution of a string generated by 2 different mechanisms (Fig.~\ref{a}A-B) and thus in 2 different regimes (random versus non-random) computer programs such as those in Figs.~\ref{a}C-D help deconvolute the string. We then do the same in all other cases but extending the number of degrees of freedom of a Turing machine tape. Notice that from our methods as illustrated in Fig.~\ref{a}A-B, the methods are invariant to direction, given that the algorithmic probability and complexity of a string and its reversal (and set of computable transformations) preserve the complexity and mechanistic origin of each object (up to a small constant which is the length of the computable transformation).

\subsection{Decomposition of Sequences and Space-time Diagrams}

\begin{figure}
\centering
\textbf{A}\hspace{6cm} \textbf{B}\\
\scalebox{.21}{\includegraphics{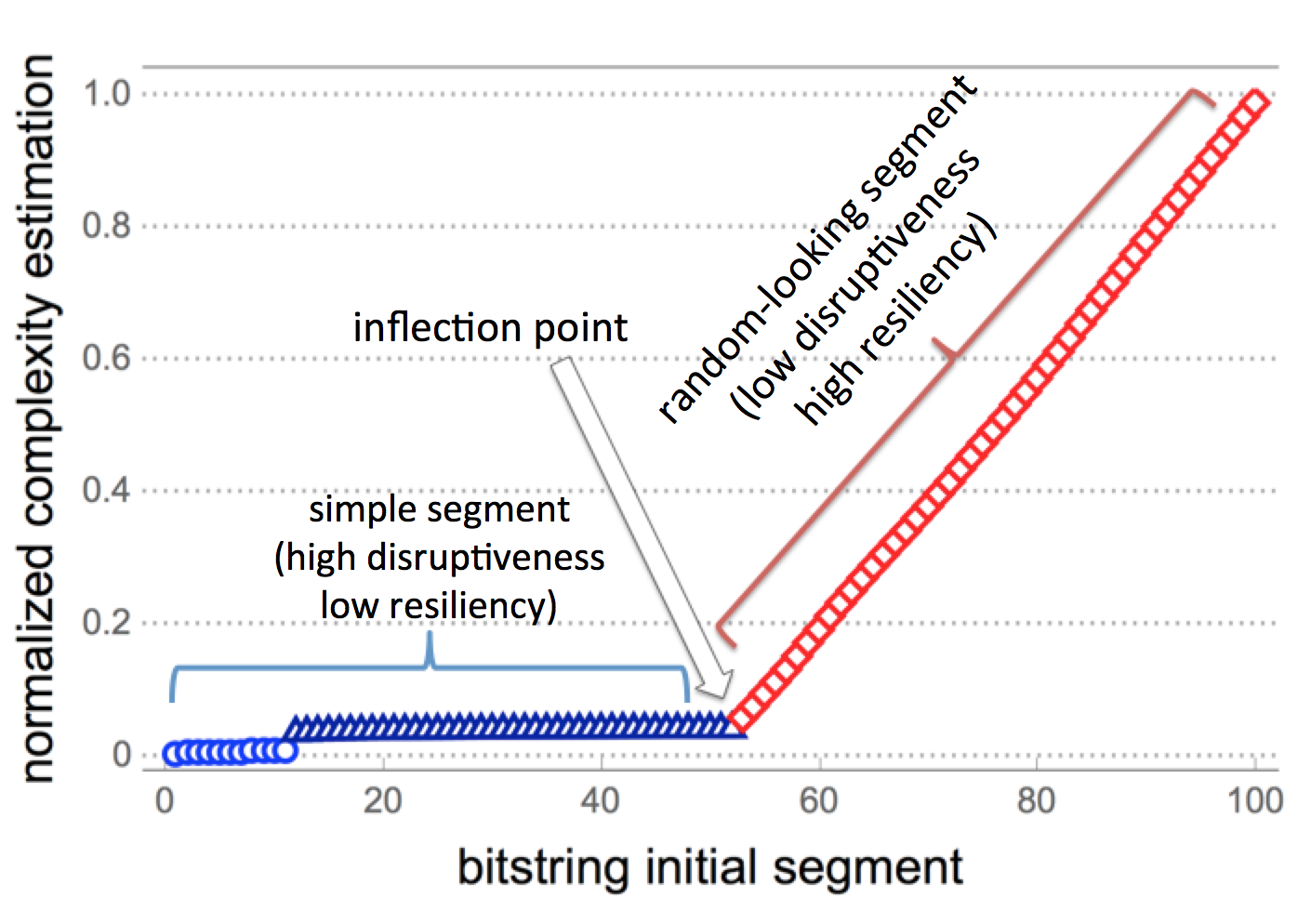}}\hspace{1cm}\scalebox{.21}{\includegraphics{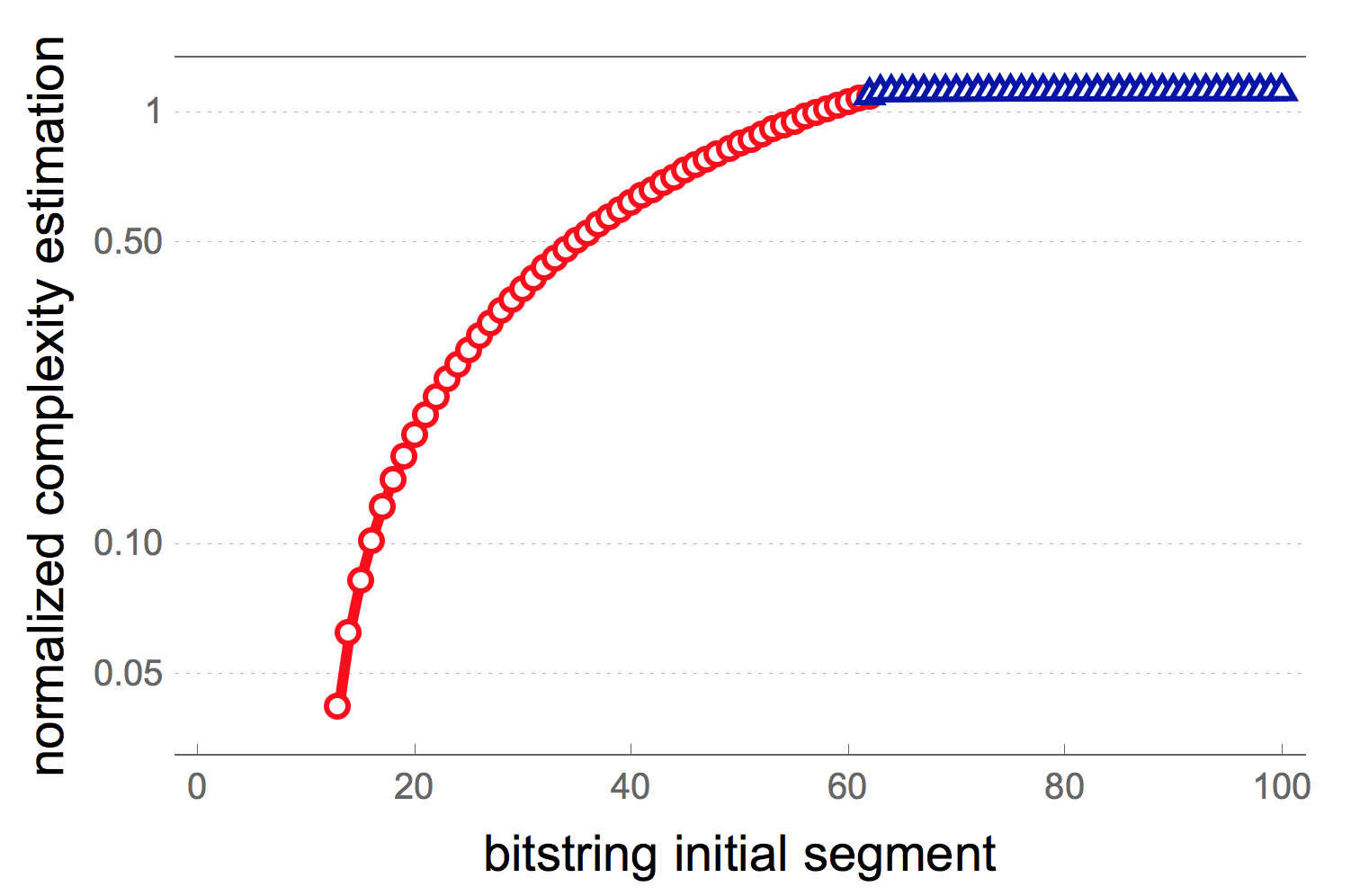}}\\

\medskip

\textbf{C}\hspace{5cm} \textbf{D}\\
\hspace{2.5cm} \scalebox{.21}{\includegraphics{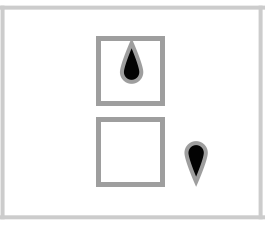}}\scalebox{.21}{\includegraphics{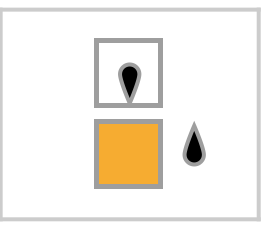}}\hspace{1.2cm}\scalebox{.26}{\includegraphics{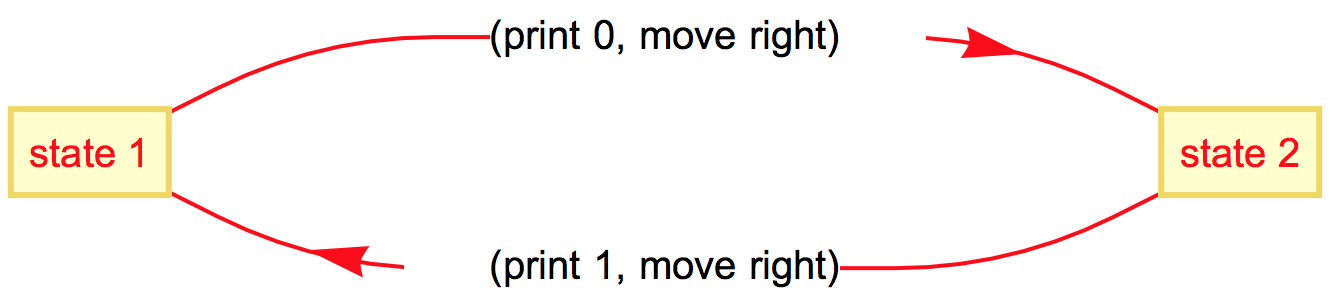}}\\

\medskip

\textbf{E}\\
\scalebox{.24}{\includegraphics{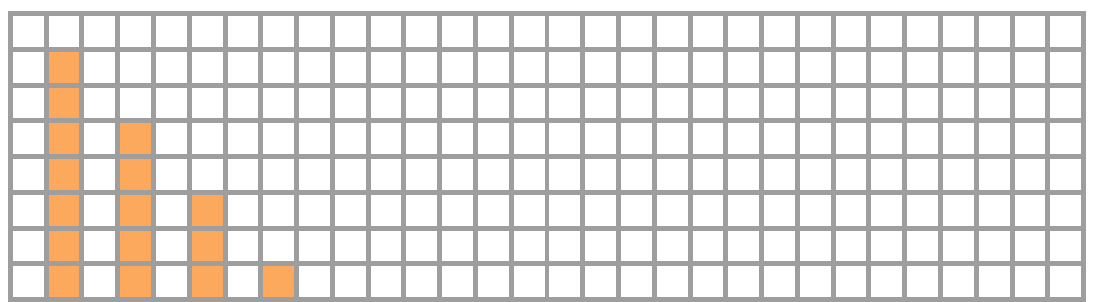}}\\

\medskip

\textbf{F}\hspace{6cm} \textbf{G}\\
\scalebox{.26}{\includegraphics{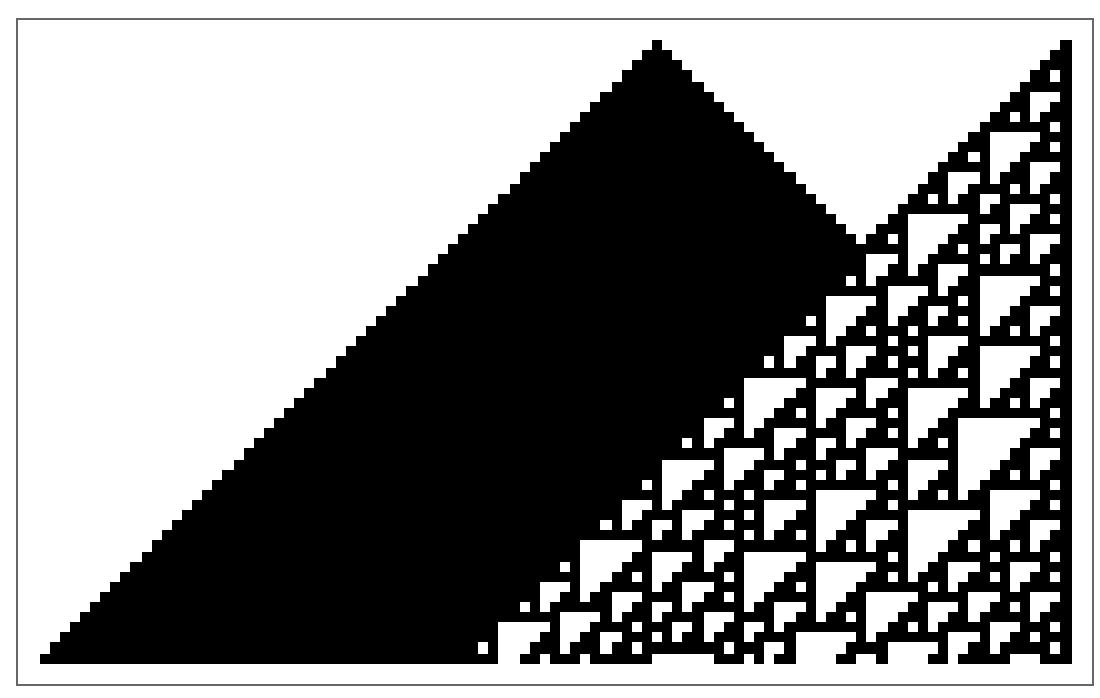}}\hspace{1cm}\scalebox{.26}{\includegraphics{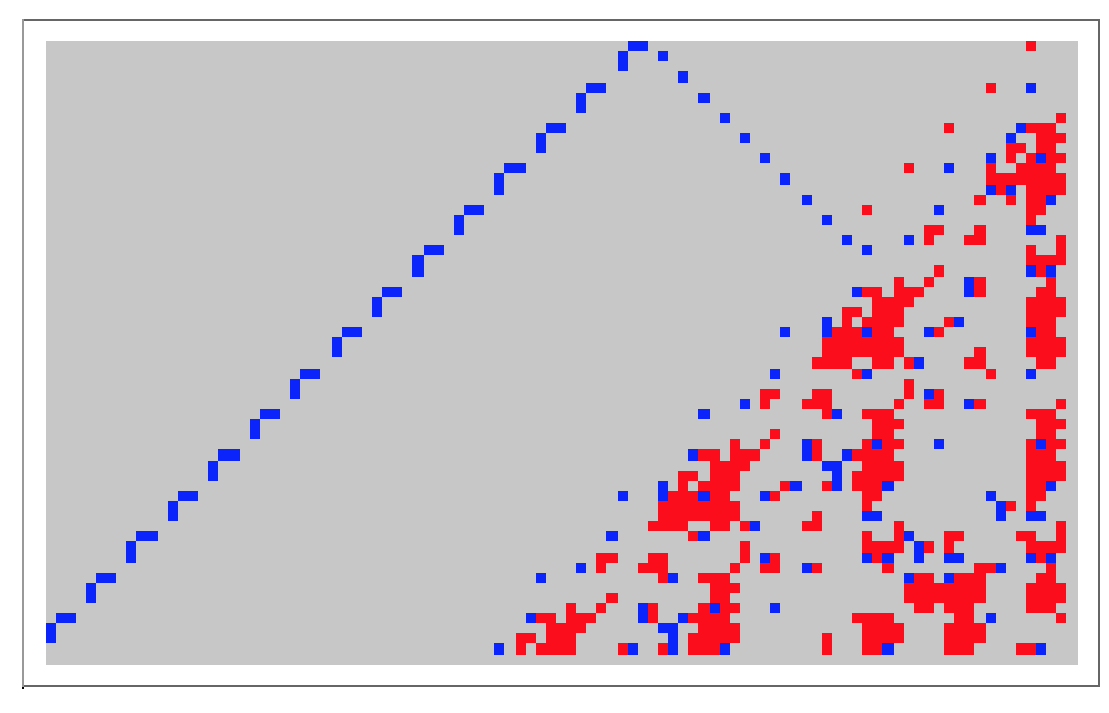}}\\

\caption{\label{a}Proof of concept applied to a binary string composed of 2 segments with different underlying generating mechanisms (computer programs). A: Log plot of complexity estimation of a regular segment (blue) consisting of the repetition of `01' 25 times followed by a random-looking segment (red). B: Log plot reversing the order of A, yet preserving the qualitative behaviour of the different segments. C: The code of the smallest generating program (a non-terminating Turing machine) depicted visually (states are arrows in different directions), producing the string of $01^n$ for any $n$ (0 is white and 1 is orange) starting from a blank tape, as shown in the space-time diagram (E). D: the same computer program as a state diagram. F: An illustration of a very simple case of interacting programs with one dominating the other and each with different generating mechanism (ECA rules 255 v 110) each running for 60 steps using interacting rule 531441 as described and explained in detail in the Sup. Inf. G: Algorithmic information footprint: every pixel is perturbed by flipping its value and evaluating its contribution to the original object coloured accordingly: If grey, it makes the no contribution, blue represents a low contribution and red high contribution (its presence contributes to its algorithmic randomness).}
\end{figure}

We tested the \textit{causal deconvolution} algorithm on different types of objects, in order to explore and explain its applicability, advantages and limitations. We start with the simplest version of an object which conveys information, a string, and move later to consider richer objects such as networks.

We will use different programs to produce different parts of a string, that is, a program $p$ to generate segment $s_1$ and a program $p^\prime$ to generate segment $s_2$ put next to each other. Clearly the string has been generated by 2 generating mechanisms ($p$ and $p^\prime$). Now we use the algorithm to deconvolve the string and find the number of generating mechanisms and most likely model mechanisms (the programs themselves) inducing a form of \textit{algorithmic partition} based on the likelihood of each segment being produced by different generating mechanisms.

Figs.~\ref{a}A-E illustrate how strings that have short generating mechanisms are significantly and consistently more sensitive to perturbations. The resulting string is 
{\color{blue}0101010101010101010101010101010101010101010101010101}{\color{red}11010010101010000000\\1001100111100110000011100110} with the colours corresponding to the parts suggested by the different regimes, according to their algorithmic contribution and the segment's resilience in the face of perturbations (by deletion and replacement) to the original string. Behind every real number approximating the algorithmic complexity of a string, there is the discovery of a large set of generating programs when using the Algorithmic Probability (AP)-based measure BDM producing the object. 

We not only could find the number of mechanisms correctly (Figs.~\ref{a}A and B) but also the candidate programs (which for this trivial example are exactly the original) that generate each segment (Figs.~\ref{a}C-E) by seeking the shortest computer programs in a bottom-up approach~\cite{d4,d5}. Finding the shortest programs is, however, secondary, because we only care about the different explanatory power that different programs have to explain the data in full or in part, pinpointing the different causal nature of the segments and helping in the deconvolution of the original observation.

Figs.~\ref{a}C-E depict the computer program (a non-terminating Turing machine) that is found when calculating the BDM of the $01^n$ string. The BDM approximation to the algorithmic complexity of any $01^n$ string is thus the number of small computer programs that are found capable of generating the same string or, conversely (via the algorithmic Coding theorem, see~\cite{bdm,kolmo2d}), the length of the shortest program producing the string. For example, the string $01^n$ was trivially found to be generated by a large number of small computer programs (in Fig.~\ref{a}C,D depicted a non-terminating Turing machine with E its output) using our algorithmic methods (as opposed to, e.g., using lossless compression, which would only obfuscate the possible generating model) with only 2 rules out of $2 \times 2$ rules for the size of Turing machine with only 2 states and 2 symbols and no more, thus of very low algorithmic complexity compared to, e.g., generating a random-looking string that would require a more complex (longer) computer program. The computer program of a truly random string will grow in proportion to the length of the random string, but for a low complexity string such as $01^n$, repeated any number of times $n$, the length of the computer program is of (almost) fixed size, growing only by $log(n)$ if the computer program is required to stop after $n$ iterations. In this case $01^n$ is a trivial example with a strong statistical regularity whose low complexity could be captured by applying Shannon entropy alone on blocks of size 2. 

\begin{figure}
\centering
\textbf{A}\hspace{5cm} \textbf{B}\\
\scalebox{.271}{\includegraphics{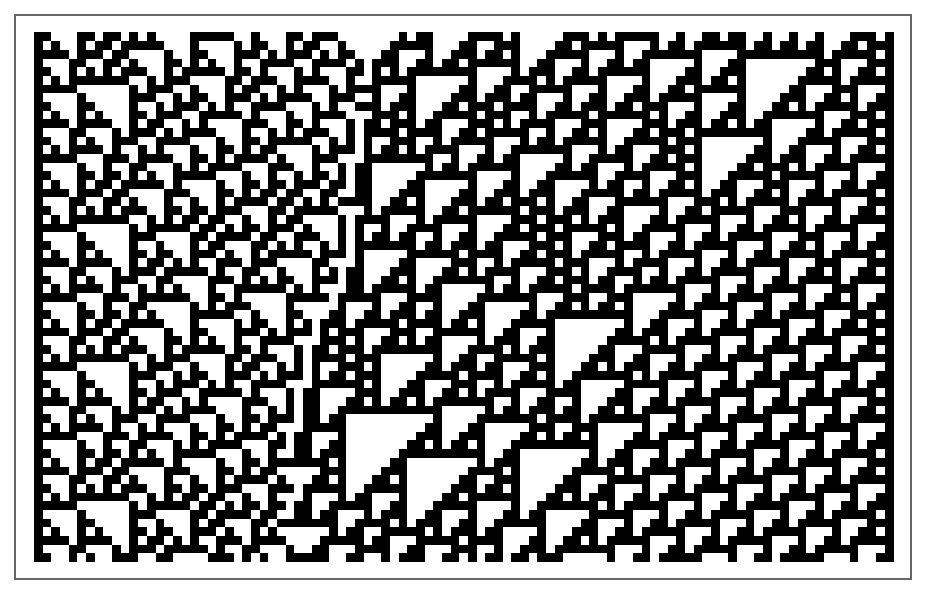}}\hspace{1cm}\scalebox{.271}{\includegraphics{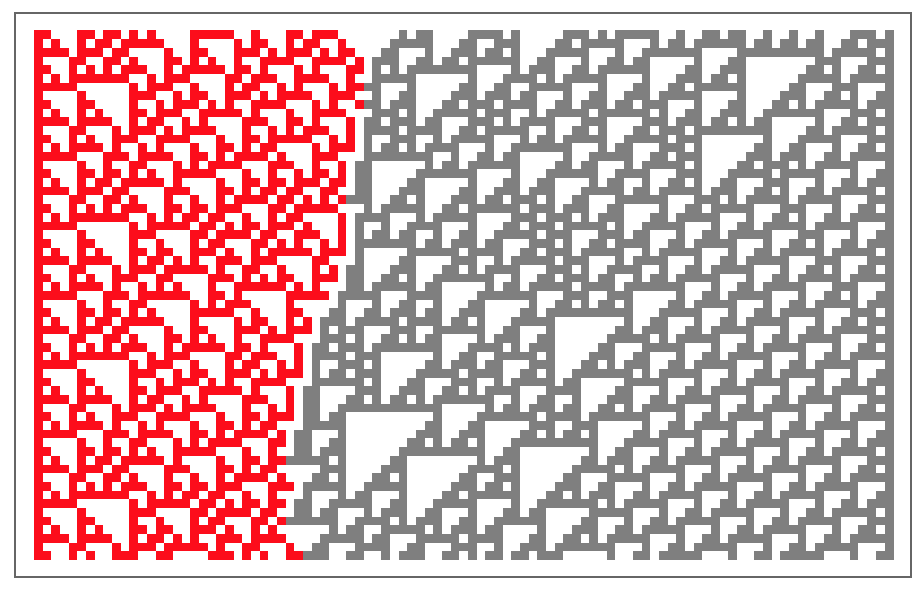}}\\

\textbf{C}\\
\scalebox{.5}{\includegraphics{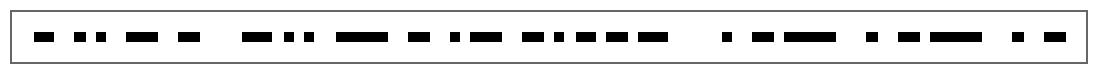}}\\

\textbf{D}\hspace{5cm} \textbf{E}\\
\scalebox{.23}{\includegraphics{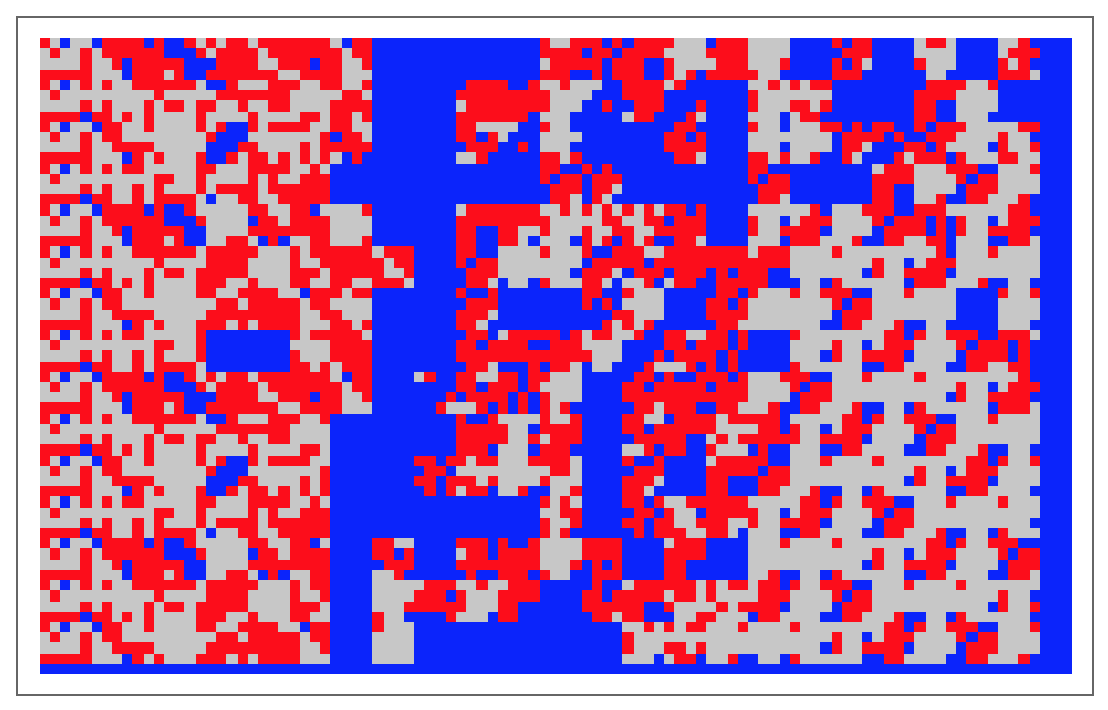}}\hspace{.8cm}\scalebox{.23}{\includegraphics{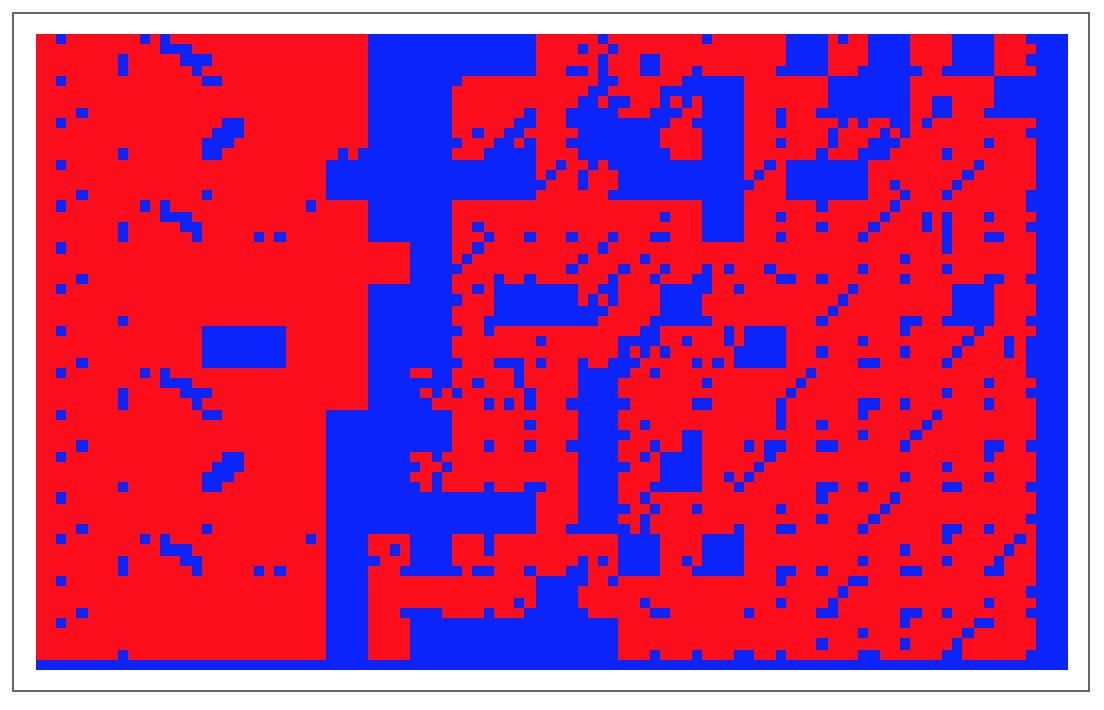}}

\textbf{F}\\
\hspace{1cm}\scalebox{.26}{\includegraphics{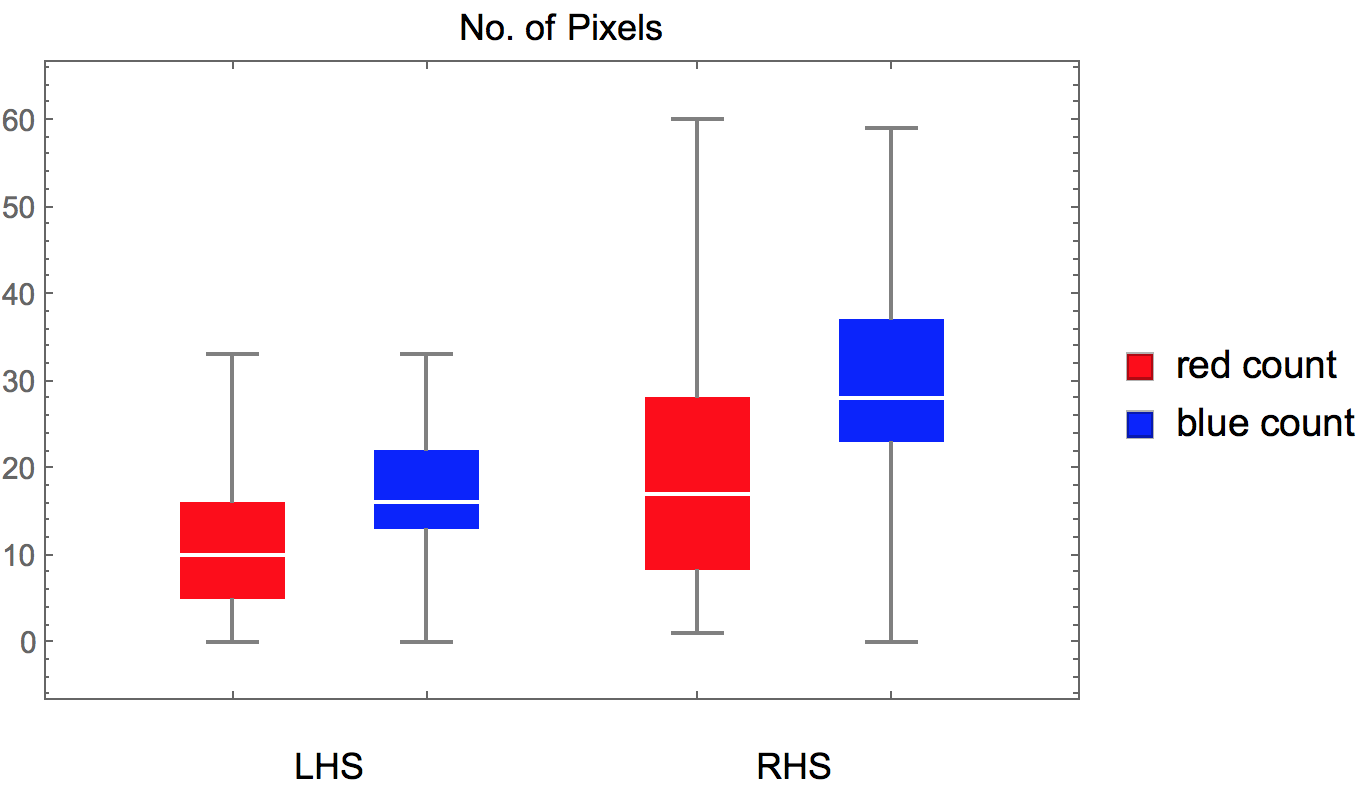}}\\

\caption{\label{b}A: The output of 2 different intertwined programs (ECA rules 60 and 110) with qualitatively complex output behaviour (11 to 60 steps depicted here, from a random initial condition) interacting with each other (one of which has been proven to be Turing-universal and the other has also been conjectured to be universal~\cite{nks}), each producing  structures of a similar type that, from an observer's perspective, are difficult to distinguish (see Subfigure C) as is artificially done in B (knowing which pixel is generated by which rule). C: What an observer of the last runtime would see in the form of a stream of bits with no clear statistical distinction. D: The algorithm pinpoints the regions of neutral, positive and negative, with the contiguous largest blue component segmenting the image into two. E: only negative vs positive causal contributions where both Shannon entropy and a popular lossless compression algorithms fail (see Sup. Inf.). F: Sanity check/validation: Statistically significant quantitative differences among the parts after application of the algorithm as illustrated in E among apparently weak qualitative differences as illustrated in Subfig. A. More cases are provided in the Sup. Inf.}
\end{figure}

Figs.~\ref{a}F-G illustrate how the algorithm can separate regions produced by generating mechanisms of different algorithmic information content by observing their space-time dynamics, thereby contributing to the deconvolution of regions that are produced by different generating mechanisms. In this example both programs are sufficiently robust to not break down (see Sup. Inf.) when they interact with each other, with rule 110 prevailing over 255. Yet, in the general case it is not always easy to tell these mechanisms apart. In more sophisticated examples, such as in Figs.~\ref{b}D-E, we see how the algorithm can break down contiguous regions separating an object into 2 major components corresponding to the different generating computer programs that are intertwined and actively interacting with each other. The experiment was repeated 20 times with programs with differing qualitative (e.g.  Wolfram class) behaviour. 

Fig.~\ref{b}F demonstrates that perturbations to regions in red have a more random effect after application and are thus by themselves less algorithmically random. When regions are of the same algorithmic complexity they are likely to be generated by similar algorithms, from algorithms that are of similar minimal length. The removal of pixels in the blue regions move the interacting system away from randomness and are themselves more algorithmically random. Blue structures on the left hand side correspond to large triangles occurring in ECA rule 110 that are usually used to compute and transfer information in the form of particles.
However, triangular patterns transfer information in a limited way because their light cone of influence reduces at the greatest possible speed of the automaton, and they are assigned an absolute neutral information value. Absolute neutral values are those closest to 0. Once separated, the 2 regions have clearly different algorithmic characteristics given by their causal perturbation sensitivity, with the right hand side being more sensitive to both random and non-random perturbations. Moreover, Fig.~\ref{b}F shows results compatible with the theoretical expectation and findings in~\cite{maininfo} where a measure of reprogrammability associated with the number and magnitude of elements that can move a dynamical system towards or away from randomness was introduced and shown to be related to fundamental properties of the attractor space of the system.

Fig.~\ref{b}C-F shows, for example, that by iterating the deconvolution algorithm not only do the 2 main components of the image correspond to the 2 generating ECA rules, but a second application of the algorithm would produce a third or more components corresponding to further resilient features generated by the rules, which can be considered rules themselves within a smaller rule (state/symbol) space. However, in the deconvolved observations the interacting rule determining how 2 or more rules may interact effectively constitutes a third global rule to which the algorithm has no direct access, or an apparent region in the observed window.

\subsection{Network Deconvolution}

\begin{figure}
\centering
\textbf{A}\hspace{5cm} \textbf{B}\\
\scalebox{.17}{\includegraphics{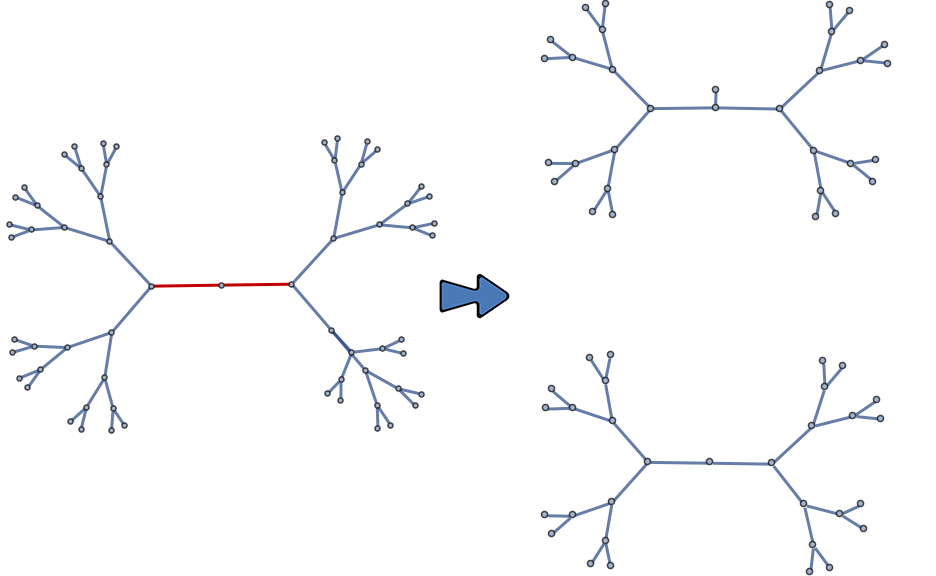}}\hspace{.5cm}\scalebox{.17}{\includegraphics{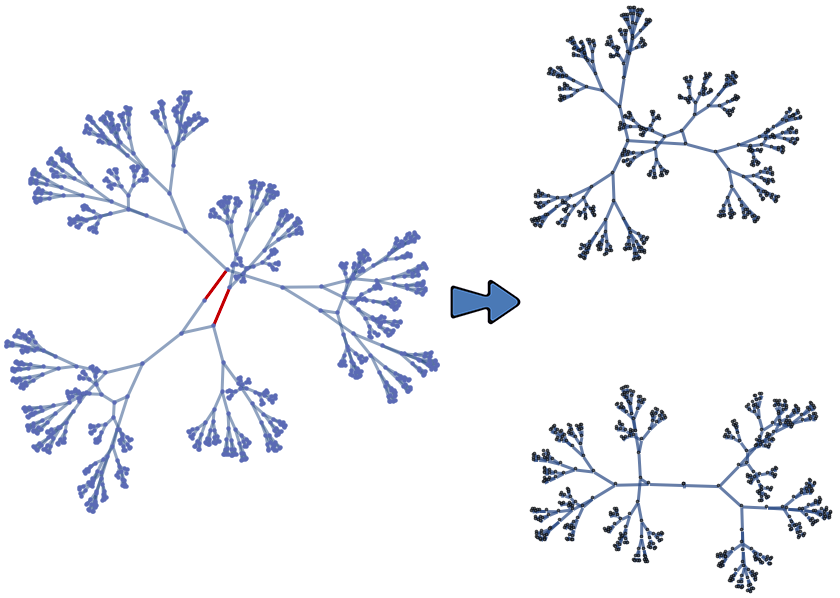}}\\

\textbf{C}\hspace{5cm} \textbf{D}\\
\scalebox{.25}{\includegraphics{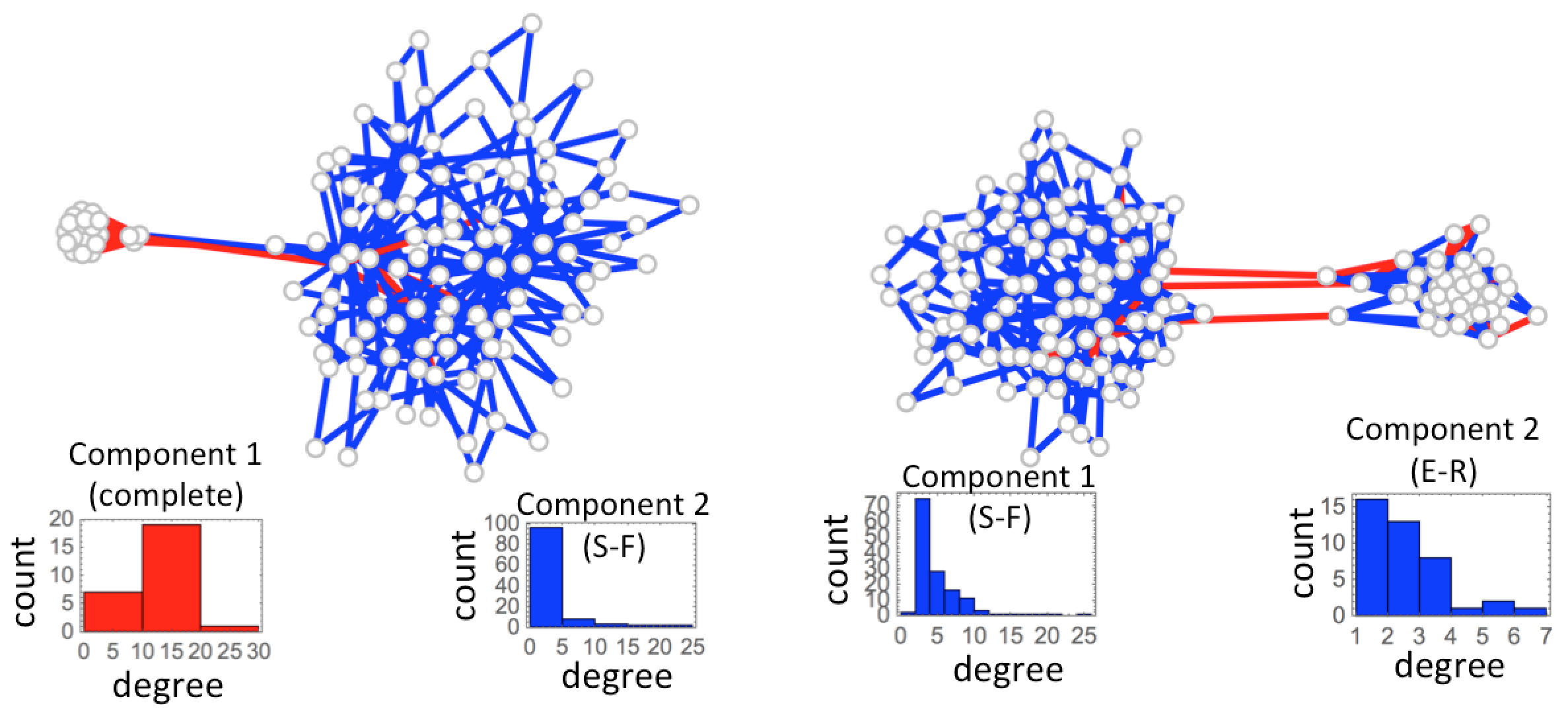}}\\

\textbf{E}\\
\scalebox{.25}{\includegraphics{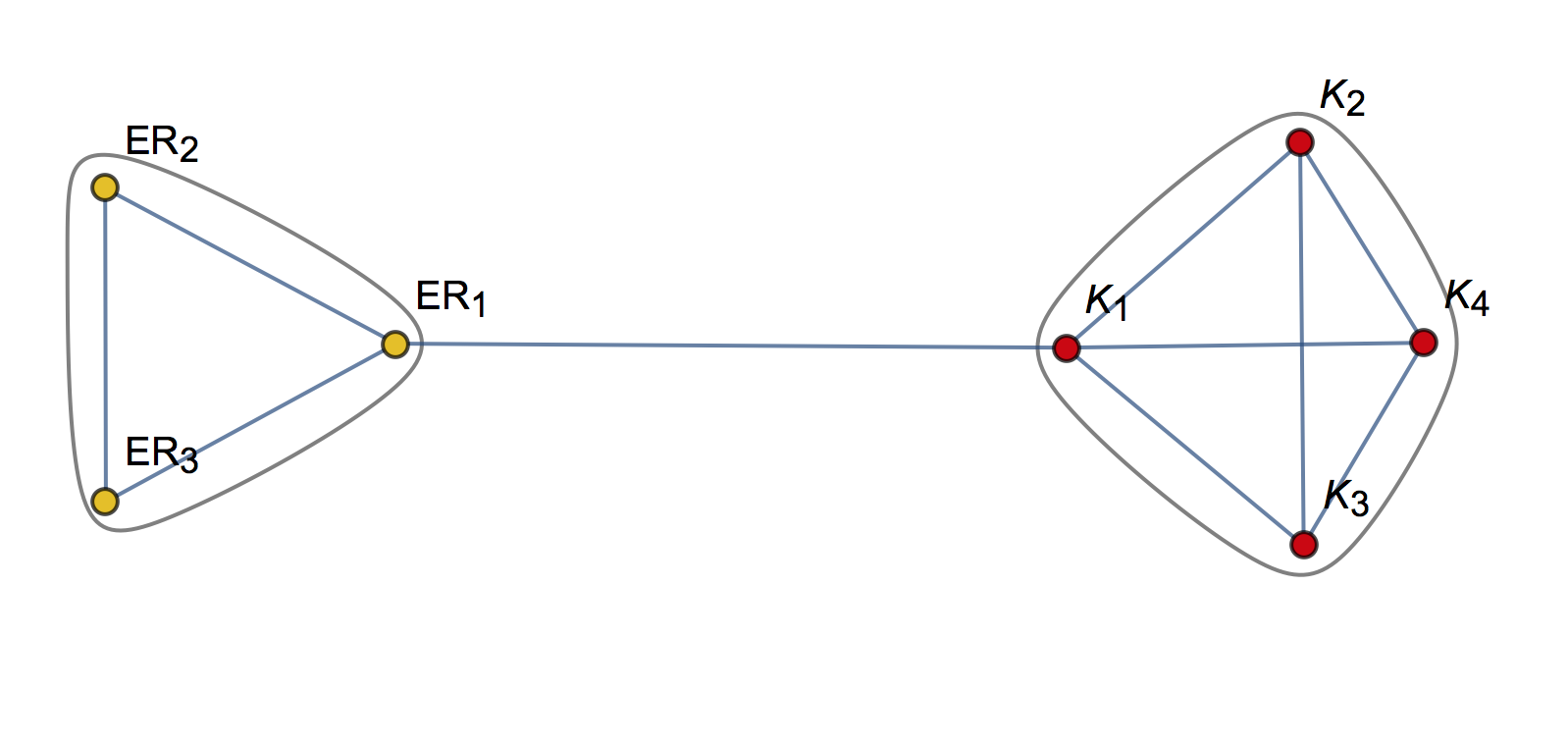}}

\caption{\label{c}A,B: Forced deconvolution of a tree by minimisation of graph algorithmic information loss thereby maximising causal resemblance of the resultant components (hence causal clustering). Depicted are the components of a $K$-ary trees of size 6 (A) and 10 (B) and their resulting graphs after one iteration of the deconvolution algorithm. C: Deconvolution of 20 cases of scale-free (S-F) networks generated by preferential attachment randomly connected to a complete graph. Negative edges break down the original graph into components corresponding to the different underlying generating mechanisms. Histograms correspond to each network according to the decomposition showing the expected degree distribution of the 2 resulting major components. D: Deconvolution of 20 cases of random graphs (E-R) connected to scale-free (S-F) networks (here depicted a typical case). E: The algorithm first separates the subcomponents with the largest algorithmic difference, followed by other subcomponents hence providing a natural hierarchy of source likelihood.}
\end{figure}

Classification can usually be viewed as solving a problem which has an underlying tree structure according to some measure of interest. One way to think of optimal classification is to discover a tree structure at some level of depth, with tree leaves closer to each other when such objects have a common or similar causal mechanism and for which no feature of interest has been selected. Fig.~\ref{c} illustrates how the algorithm may partition data, in this case starting from a trivial example that breaks complete $K$-ary trees. Traditionally, partitioning is induced by an arbitrary distance measure of interest that determines the connections in a tree, with elements closer to a cluster centre connected by edges. The algorithm breaks the trees (see Fig.~\ref{c}) into as many components as desired by iterating over remaining elements if required until the number of desired components is obtained or the terminating criterion is applied (c.f. Subsection~\ref{terminating}). Figs.~\ref{c}A,B provide examples illustrating how to maximise topological symmetry. The algorithm can be applied, without loss of generalisation, to any non-trivial graph, as in Figs.~\ref{c}C,D or on any dataset for that matter.

\begin{figure}[htp]
\centering
\textbf{A}\hspace{5.3cm} \textbf{B}\\

\bigskip

\scalebox{.37}{\includegraphics{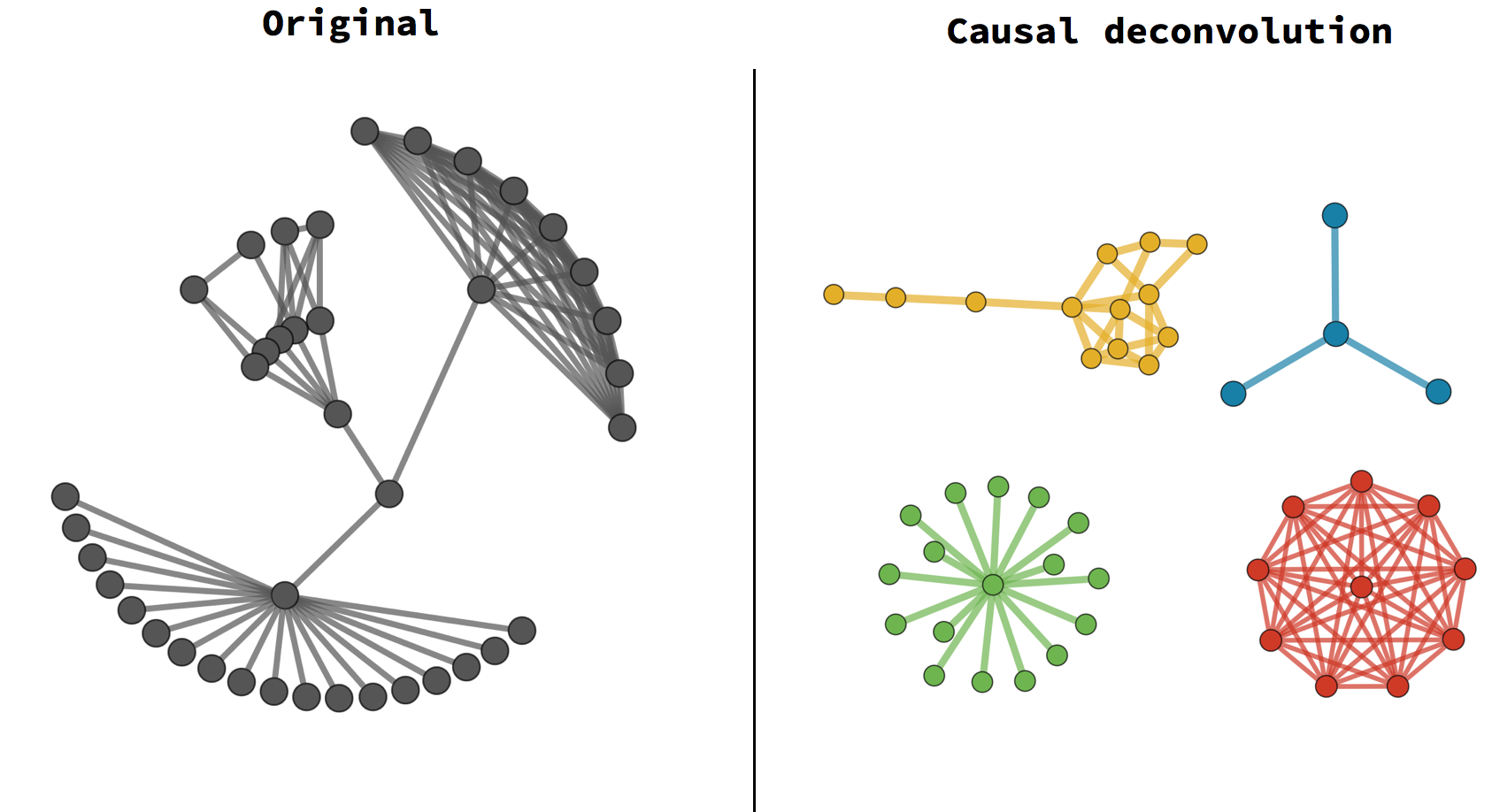}}\\

\bigskip

\textbf{C}\hspace{5cm} \textbf{D}\\

\bigskip

\scalebox{.287}{\includegraphics{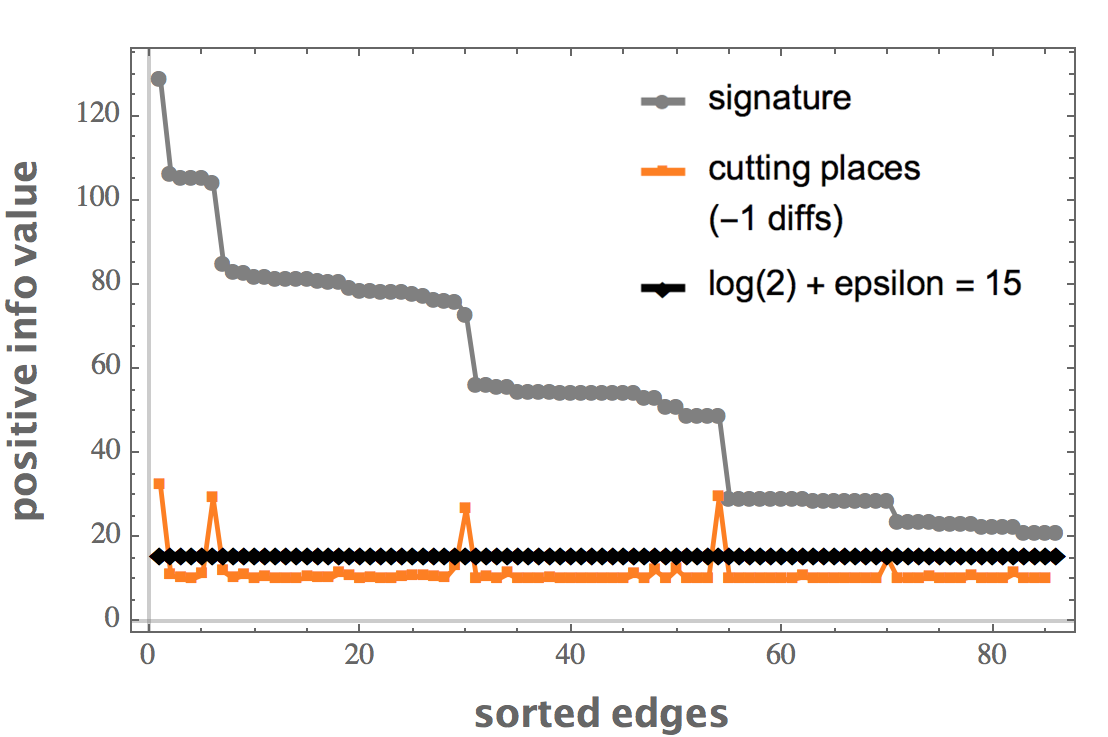}}\hspace{.8cm}\scalebox{.277}{\includegraphics{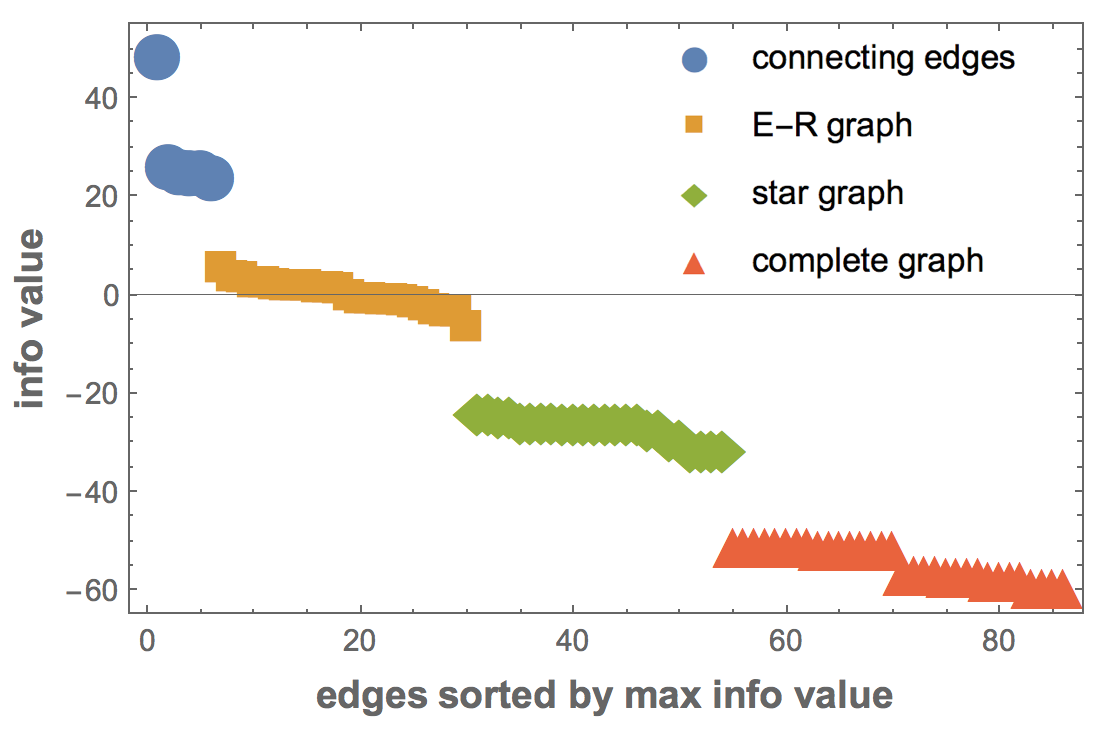}}\\
\caption{\label{c2}A: Convoluted graph composed of 3 subgraphs produced by different generating mechanisms. B: Deconvolution of (A) by Algorithm 2, keeping edges connected if their removal does not produce a change in the algorithmic complexity of the original graph larger than $\log(2) + \varepsilon$. C: The \textit{information signature} (red line with circle markers) illustrates the distribution of information values for each edge ($x$-axis) in the original graph (A). Also shown is a line of the differences of consecutive values of the signature  multiplied by -1 (blue line with square markers), indicating the breaking points (the peaks that mark the edges to be deleted) with, in this case, four peaks/values clearly standing out beyond the $\log(2)+\varepsilon$ line (orange rhombus) breaking the signature corresponding to each subgraph forming with high accuracy, thereby deconvolving the original graph (A) into the subgraphs (largest components) that are most likely generated by the same causal/algorithmic source. D: Signatures decomposition according to the breaking points found in (C) giving the colours to the subgraphs in (B) with results matching the theoretical expectation.}
\end{figure} 

Figs.~\ref{c2} illustrate the algorithm and terminating criterion starting from an artificial graph composed of several graphs (2 simple and one E-R random: a small E-R random graph connected to a star graph and to a complete graph). The graph can be successfully decomposed by algorithmic probability (see Figs.~\ref{c2}B and D) by identifying the likelihood of an edge being produced by the same mechanism by virtue of being close to each other in the information contribution (which theoretically should be removed by only $\log(2)$ if it follows the normal evolution of the same process), hence what we call causal separation/partition and clustering. Fig.~\ref{c2}D with the broken components that were found above $\log(2)+\varepsilon$, also showing the distribution of edges coloured by graph membership perfectly corresponding to the different subgraphs that were used to compose the original graph Fig.~\ref{c2}A.

The same task using classical information theory (Shannon entropy) is shown not to be sensitive enough (see Sup. Inf.), and a popular lossless compression algorithm (Compress based on LZW) provided a noisy approximation (see Sup. Inf.) to the results obtained by using the Block Decomposition Method, as defined in~\cite{bdm}, whose description is provided in the Sup. Inf. 

Figs.~\ref{c}C-E illustrate how randomly connected graphs with different topologies can be broken into their respective generative mechanisms. Fig.~\ref{c}C is a complete graph of size 20 randomly connected by 3 edges to a scale-free graph of size 100. The graphs are generated by different mechanisms. One is a small program that, given a number $N$ of nodes, produces a graph with all nodes connected to all other $N-1$ nodes and has a program of short length that grows only by $\log N$~\cite{zenilkianitegner}. The scale-free network is generated by the canonical preferential attachment algorithm with 2 edges per node and requires a slightly longer algorithm that grows by $\log N + c$~\cite{zenilkianitegner} where $c$ is a small constant accounting for the pseudo-random choice of attachment nodes. The algorithm breaks the graphs into 2 components, each of which corresponds to the graphs with different degree distribution (depicted below each case) associated with its generating mechanism. This is because $|P(G_1)| + |P(G_2)| + \ldots + |P(G_n)| + |P(e_{G_i})| > |P(G_1 G_2\ldots G_n)|$ for any $G_i$, where $e_{G_i}$ is the set of edges randomly connecting $G_i$ to $G_j$ for any $i$ and $j$ for all $G$ of low algorithmic complexity.

\subsection{Robustness and Limitations}

\begin{figure}[ht!]
\centering
\textbf{A}\hspace{2.5cm} \textbf{ }\\
\medskip

\scalebox{.27}{\includegraphics{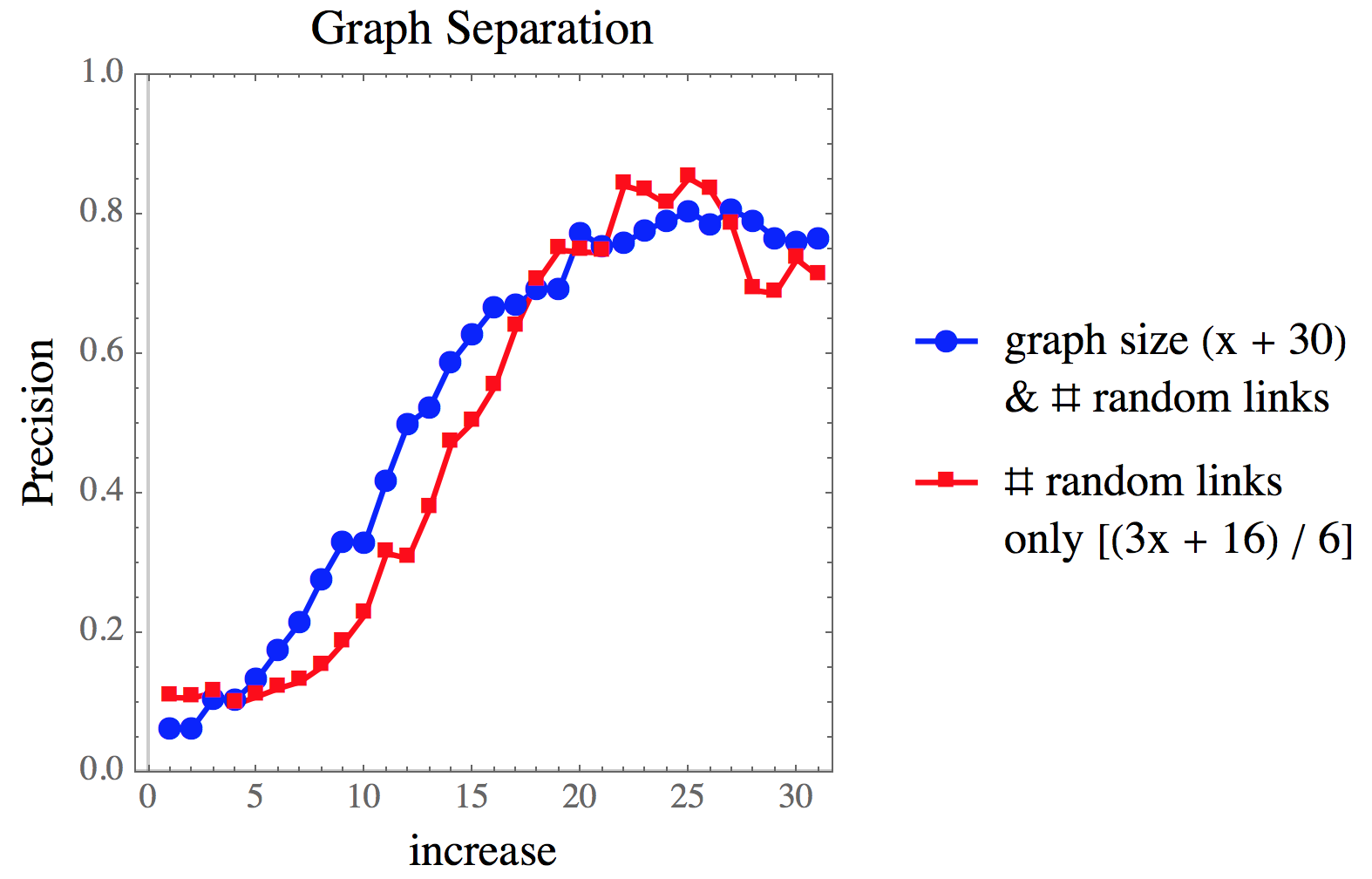}}\\

\textbf{B}\hspace{6.5cm}\textbf{C}\\
\scalebox{.3}{\includegraphics{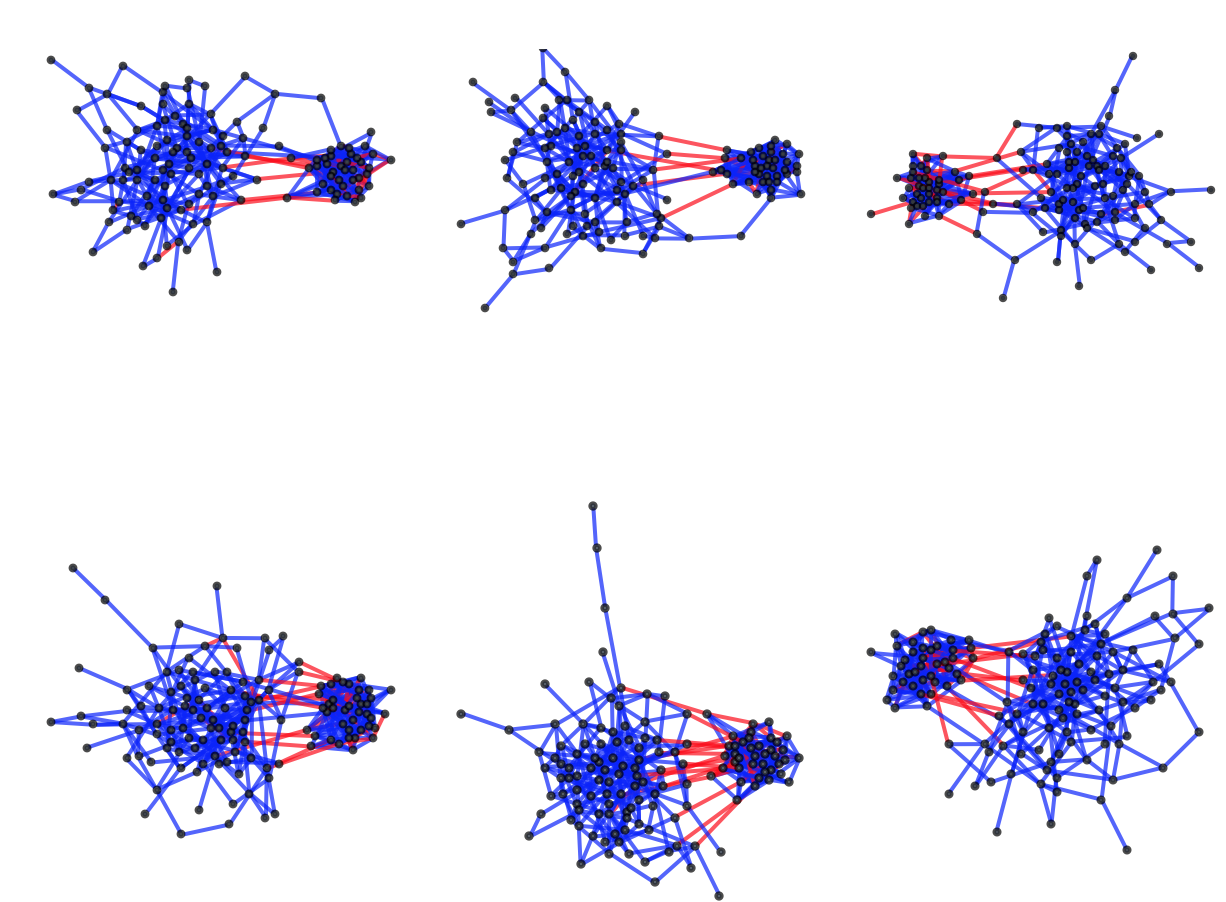}}\hspace{.7cm}\scalebox{.17}{\includegraphics{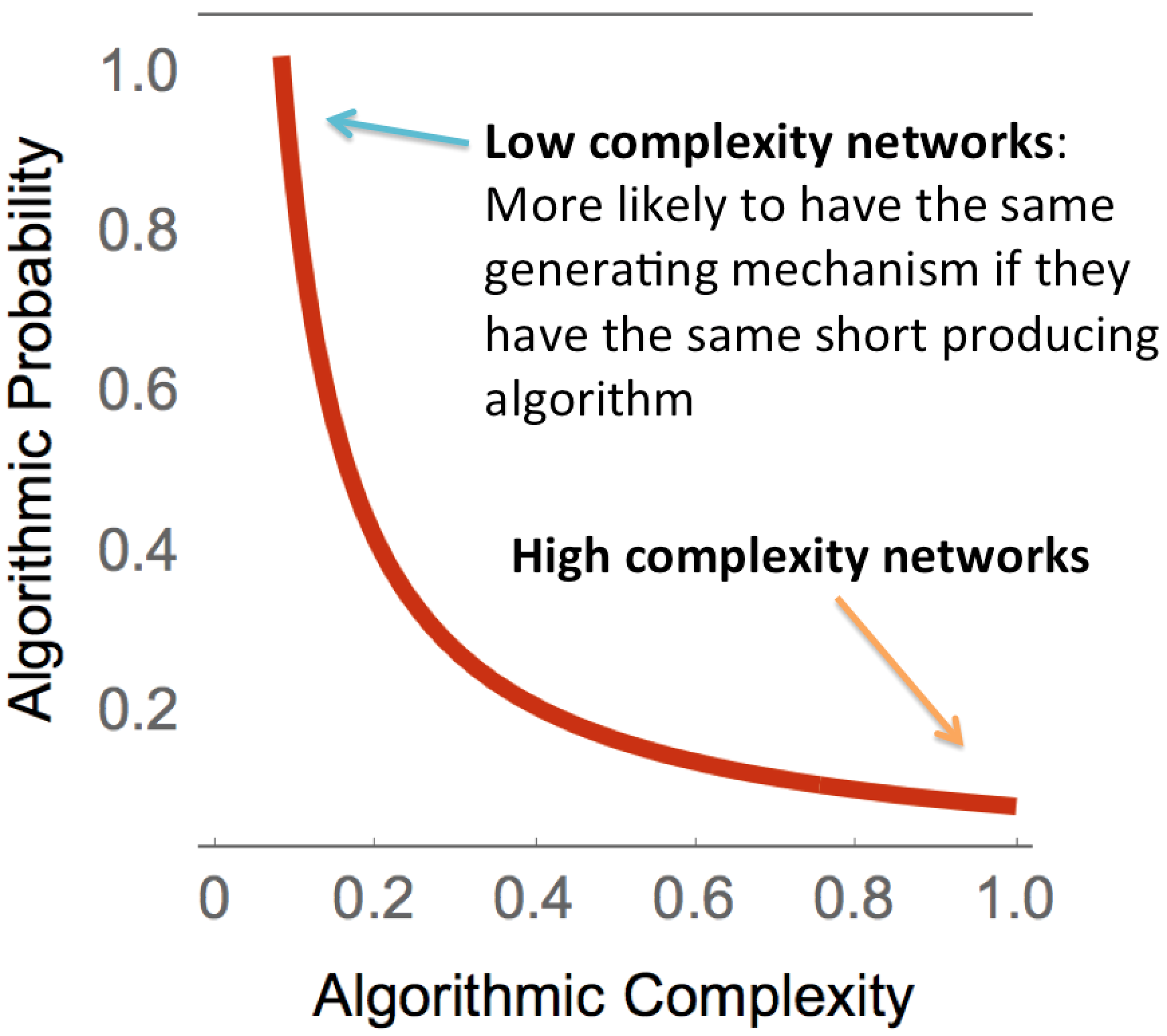}}

\caption{\label{x}A: Deconvolving a S-F graph generated by a preferential attachment algorithmic (2 new links per added node) from an E-R (0.5 edge density) graph emulating noise (each point represents the average of 10 replicates). When the number of links increases as a function of the subgraph sizes, the separability is robust (red square markers) compared to increasing the number of random links only for graphs of fixed size (40 nodes), when successful separation is compromised, noise (E-R) and structure (S-F) being indistinguishable. B: A small sample of 6 graphs among the $10 \times 20 = 200$ graphs of S-F connected to E-R graphs considered in this study. Success at identifying randomly connecting links is shown and was found to be very robust. C: Objects have exponentially greater chances of being produced by the same generating mechanism if they are of low algorithmic randomness and thus of high algorithmic probability.}
\end{figure}

Fig.~\ref{c}D illustrates a similar case to Fig.~\ref{c}C, but instead of a complete graph an Erd\H{o}s-R\'enyi (E-R) graph with edge density 0.5 is produced and connected by 3 random edges to a scale-free network produced in the same fashion as in Fig.~\ref{c}C. Again, the algorithm was able to break it down into the 2 corresponding subgraphs. Fig.~\ref{c}D represents a test case to evaluate the effect of additive noise by connecting an E-R graph of increasing size and with an increasingly greater number of random edges.

Next we ask how much structure, if any, can be recovered/extracted when adding a random (E-R) graph to different types of structured networks. To this end, we conducted a series of numerical experiments shedding light on the limitations of the algorithm introduced here in the face of additive noise. The same results were obtained for the simpler case of connecting any complete graph of increasing size to any other, such as an E-R or S-F graph.

Fig.~\ref{x} shows the results from the experiments separating graphs, in this case a scale-free graph (S-F) from an Erd\H{o}s-R\'enyi graph (E-R), the former generated by a Bar\'abasi-Albert preferential attachment algorithm~\cite{barabasi} and the latter produced by a pseudo-random generator. Fig.~\ref{x}A quantifies the error and optimal signal-to-noise ratio for optimal deconvolution, testing the algorithm under additive noise both for fixed and growing size subcomponents. Fig.~\ref{x}B shows links coloured in red as identified by the algorithm having the highest algorithmic information value when their removal sends the original composed system towards lower algorithmic information content, thereby telling apart the 2 subcomponents. Negative links are mostly on the side of the E-R graph. In other words, the S-F can be extracted with the greatest precision and a lower rate of false positives from the mix, which is to be expected given the random nature of the added links that connect the graphs, making them more like the E-R links than the S-F. If only the number of random links among graphs increases for fixed size graphs (Fig.~\ref{x}B blue circle marks), a maximum precision of about 0.9 is reached before degradation. That is, at around 32.5\% of the links randomly connecting the components. In other words, the algorithm is robust, telling apart noise from structure even after up to $0.325 \textnormal{ (from the random links connecting the components)} + 0.5 \textnormal{ (from the}\\ \textnormal{E-R component)} = 0.825$, i.e. 82.5\% of all links are random. On the other hand, the number of false positives is constant at about 5\%. In the case shown in Fig.~\ref{x}B all of the false positives (red links not connecting the 2 graphs with different topologies) are inside the E-R graph and mostly nonexistent on the side of the less random S-F graph.

\section{Conclusions}

We have introduced a new conceptual framework involving parameter-free methods as a contribution to tackling the problem of information decomposition by means of \textit{causal deconvolution} based on the fundamental theory of algorithmic probability which defines optimal inductive inference. Our approach enables a general-purpose method that removes the need for pre-defined user-centric definitions of data features such as distance metrics or associated probability distributions. We have demonstrated that the algorithm is also sufficiently robust to disentangle sophisticated intertwined systems by their likelihood of their generating mechanisms based on their algorithmic probability. 

These methods are different in nature from those used in other approaches, even those based on popular lossless compression algorithms to estimate algorithmic complexity, and in particular, those from classical information theory and of other statistical tradition. Comparisons to other methods and measures also indicate that the approach is both accurate and sensitive even when fast implementations (single pixel perturbations) are followed (as opposed to e.g. full subset perturbations) meaning that results can further be improved by expanding the set of perturbations but at the same time the algorithm is robust.

Current approaches in machine and deep learning are ill-equipped to deal with aspects of inductive inference, explanation and causation. We think that these methods open the possibility of parlaying novel mechanistic methods into more elaborate machine learning approaches with complementary and better equipped approaches based on algorithmic first principles. Our methods promote the use of techniques from causal and counterfactual analysis to tackle challenges of causal discovery based upon and complemented by universal principles drawn from the theory of computability and algorithmic complexity to take the full step towards a framework removed from traditional statistics and classical probability.

\section*{Contributions}

H.Z., N.A.K. and J.T. conceived and designed the algorithms. H.Z. designed the experiments and carried out the calculations and numerical experiments. A.A.Z. and H.Z. conceived the online tool to illustrate the method applied to simple examples and based on this paper. All authors contributed to the writing of the paper.

\section*{Acknowledgements}

H.Z. was supported by Swedish Research Council (Vetenskapsr\r{a}det) grant No. 2015-05299.

\newpage

\section*{Supplementary Information}

\subsection{2-System Interactions}

The qualitative behaviour of each program can heuristically be identified by what is known as its Wolfram class, which in turn has been formalised using tools and methods from algorithmic complexity in \cite{zenilca,zenilchaos}. Informally, Wolfram class 1 represents evolutions of programs that converge to a simple fixed configuration, exemplars of Wolfram class 2 converge to repetitive simple behaviour, those of  Wolfram class 3 produce unbounded, apparently statistically random behaviour, and exemplars of Wolfram class 4 reproduce apparently open-ended persistent structures. None of what has been introduced here depends on this behavioural characterisation based on different heuristics, and it is thus in no way fundamental to the results reported.

\begin{figure}[ht]
\centering
\textbf{A}\hspace{6.5cm} \textbf{B}\\
\scalebox{.218}{\includegraphics{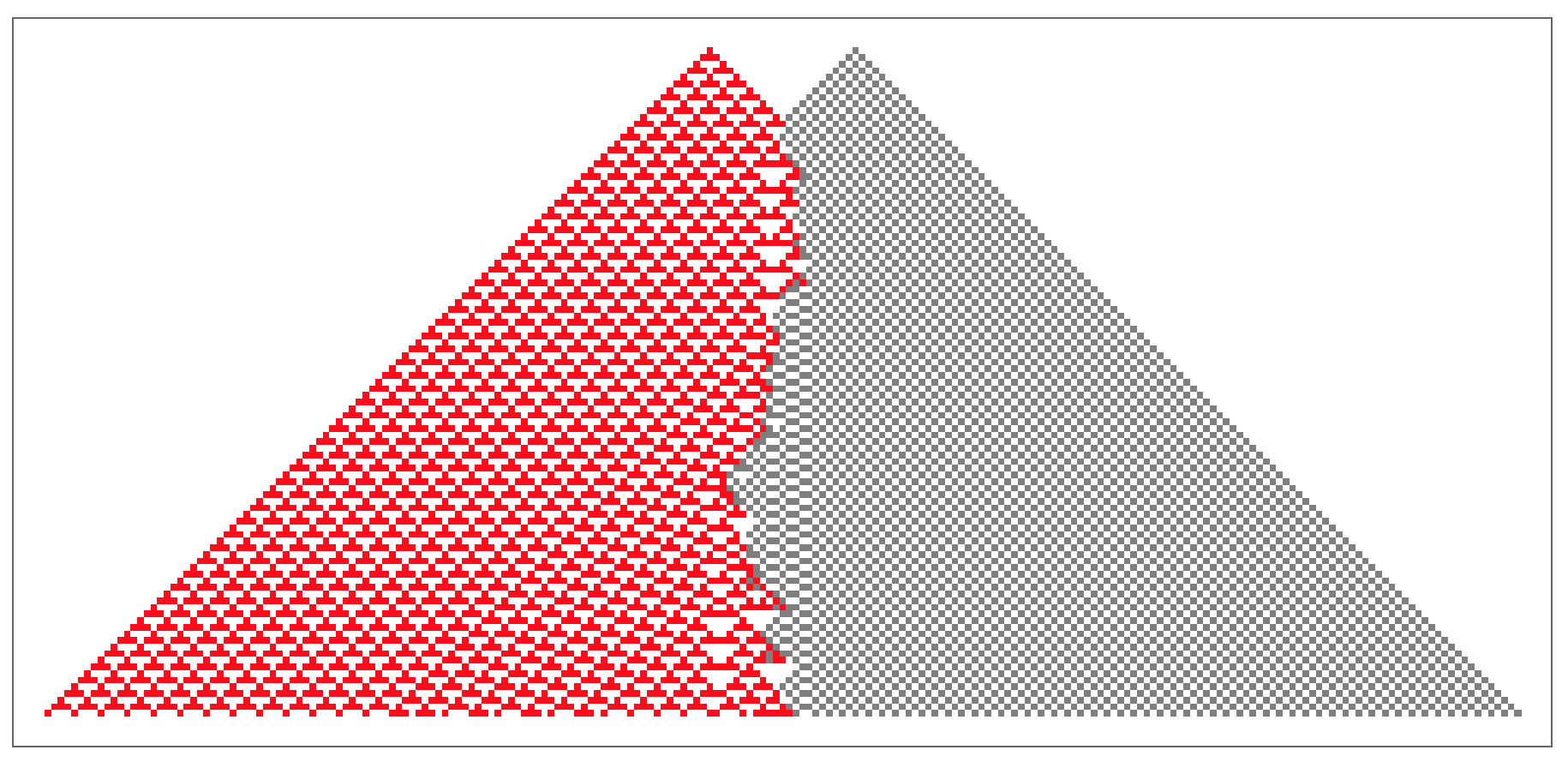}}\scalebox{.22}{\includegraphics{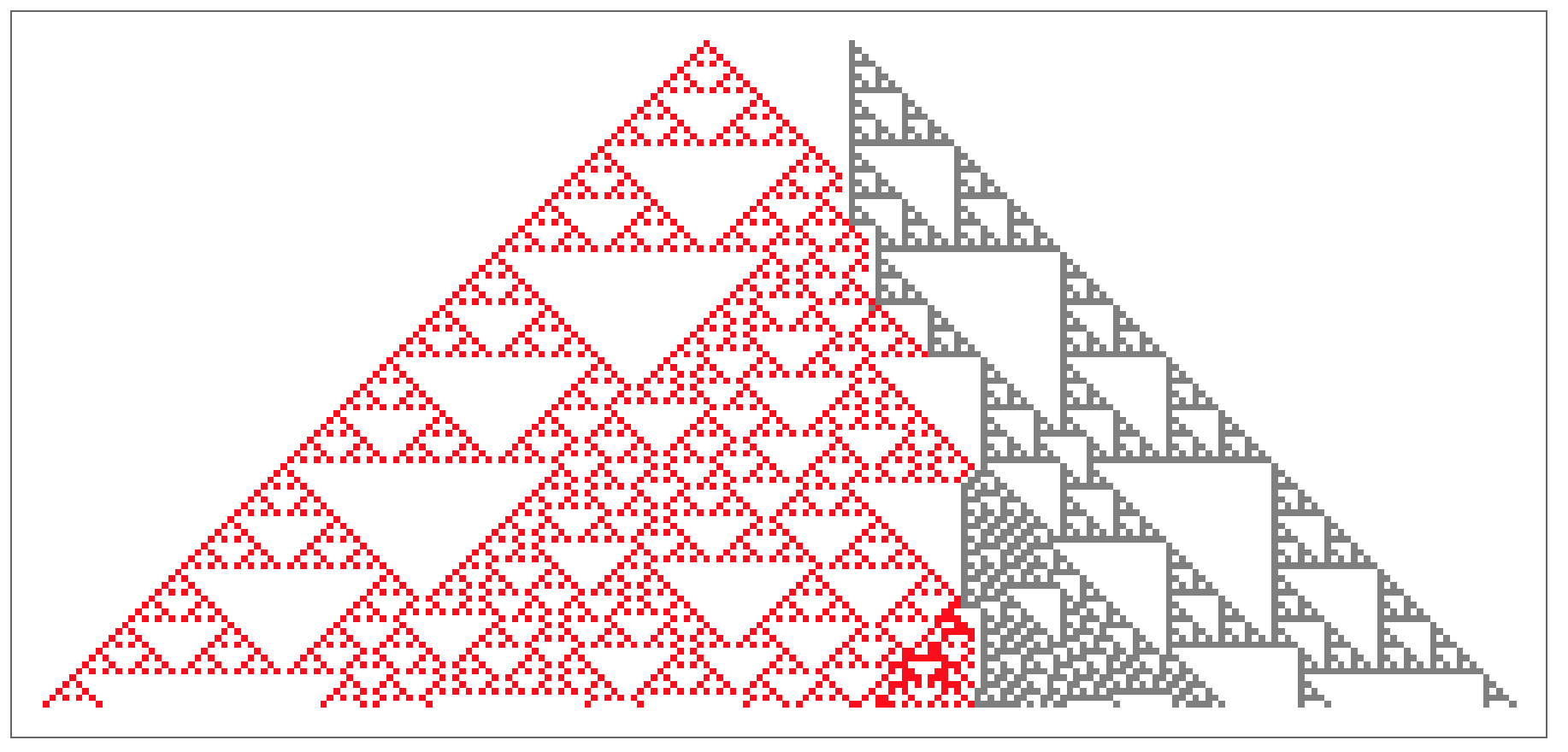}}\\
\bigskip

\textbf{C}\\
\scalebox{.28}{\includegraphics{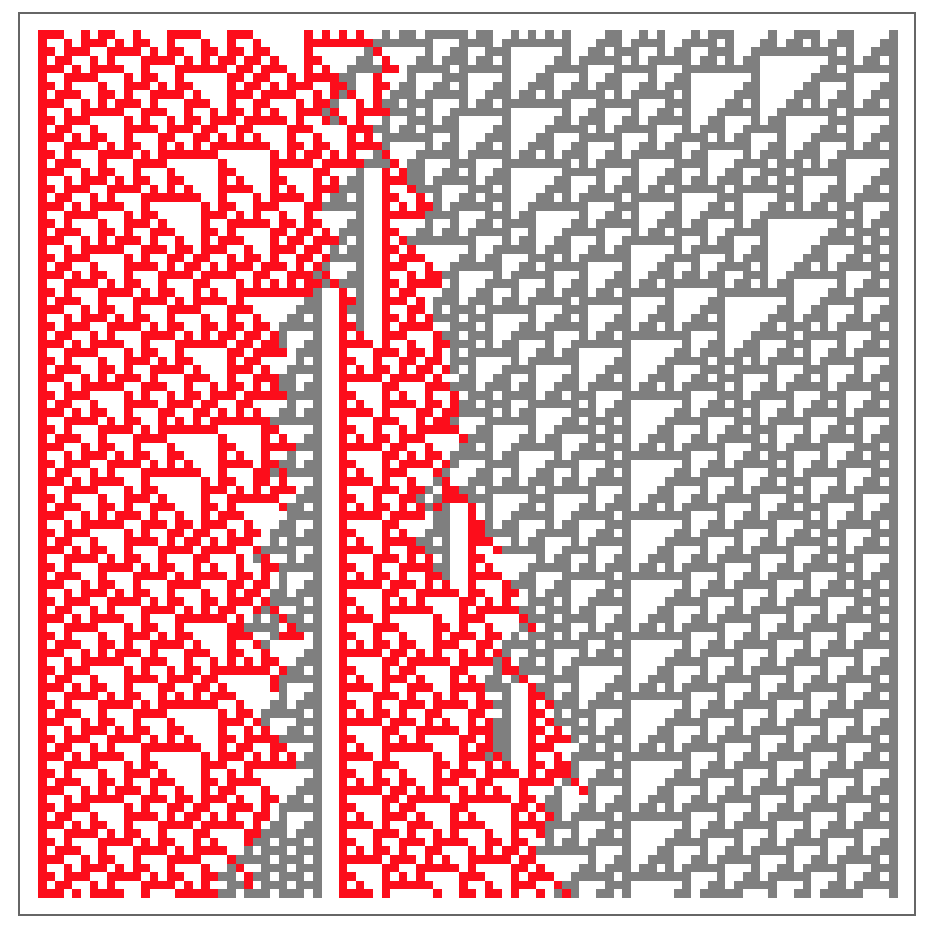}}

\caption{\label{si1}Examples of 2-system interactions using Elementary Cellular Automata (ECA). A: rules 54 (left) and 50 (right). B: rules 82 (left) and 110 (right). C: rules 60 (left) and 110 (right). Each running for 100 steps each. In all cases left and right rules can be seen to `spill' into each others space-time with non-trivial interacting rules.  In all cases, the rule governing the interaction of the 2 ECAs is rule with number 531441 according to the code provided in this Sup. Inf.}
\end{figure}

\begin{figure}[ht]
\centering

\textbf{A}\hspace{6.5cm} \textbf{B}\\
\scalebox{.22}{\includegraphics{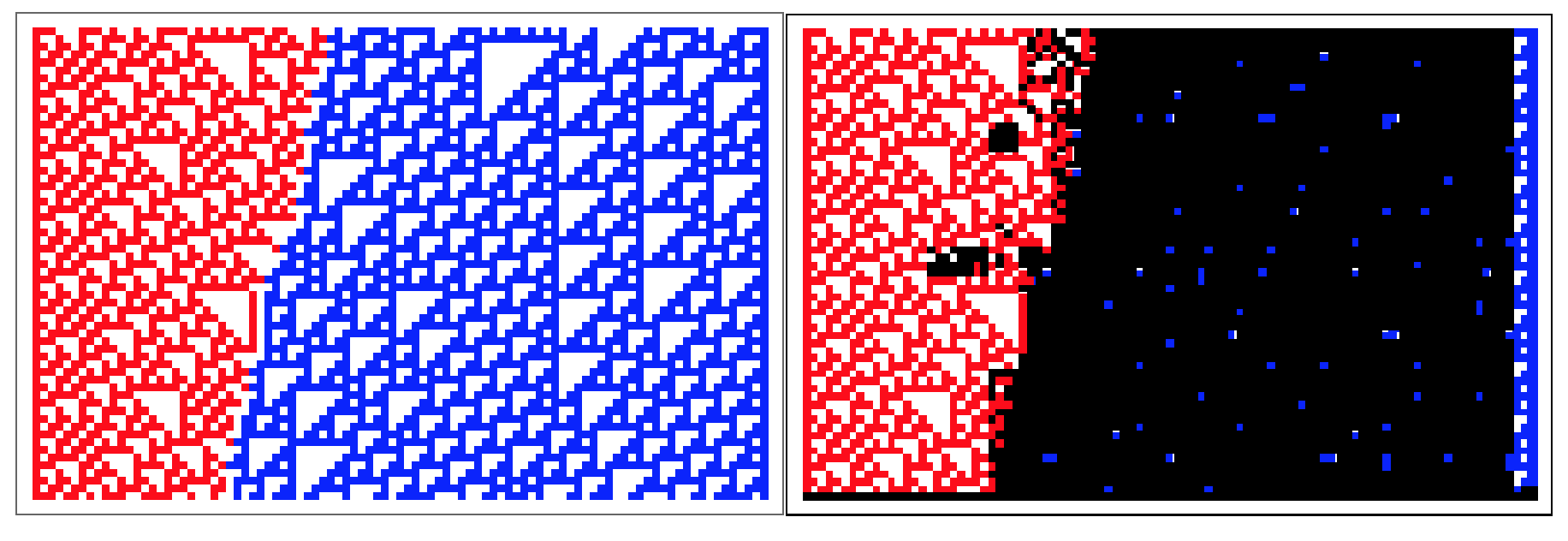}}\scalebox{.22}{\includegraphics{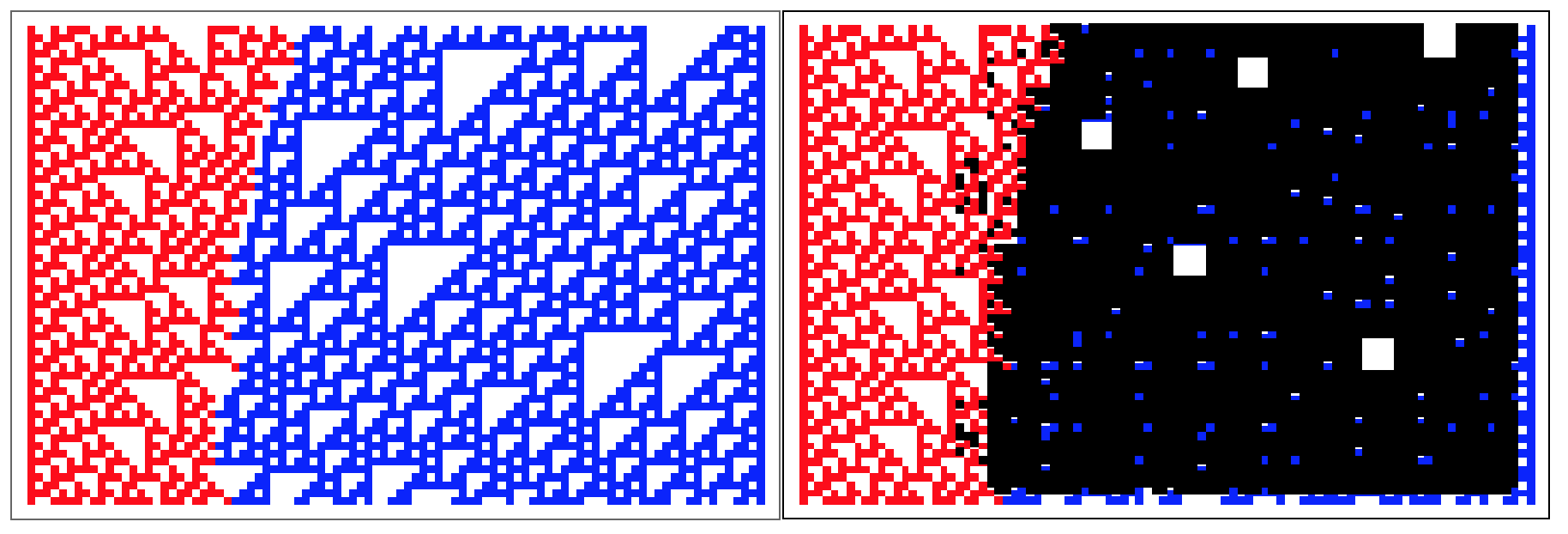}}\\
\medskip

\textbf{C}\\
\scalebox{.22}{\includegraphics{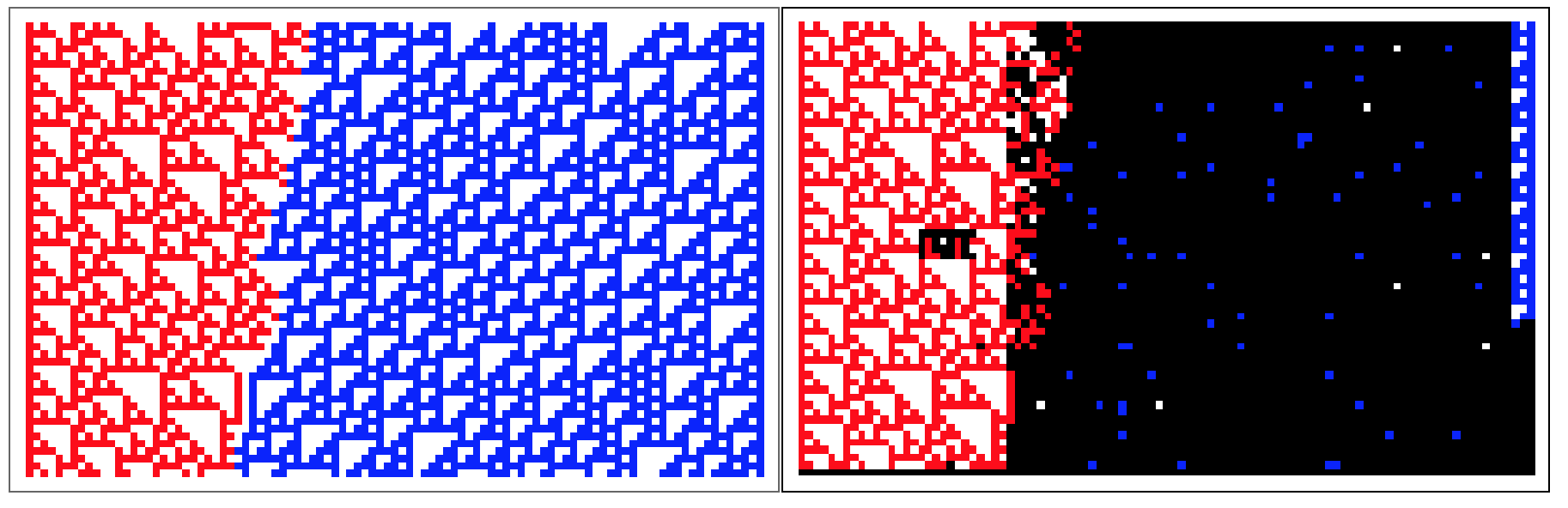}}\\
\medskip

\textbf{D}\hspace{6.5cm} \textbf{E}\\
\scalebox{.21}{\includegraphics{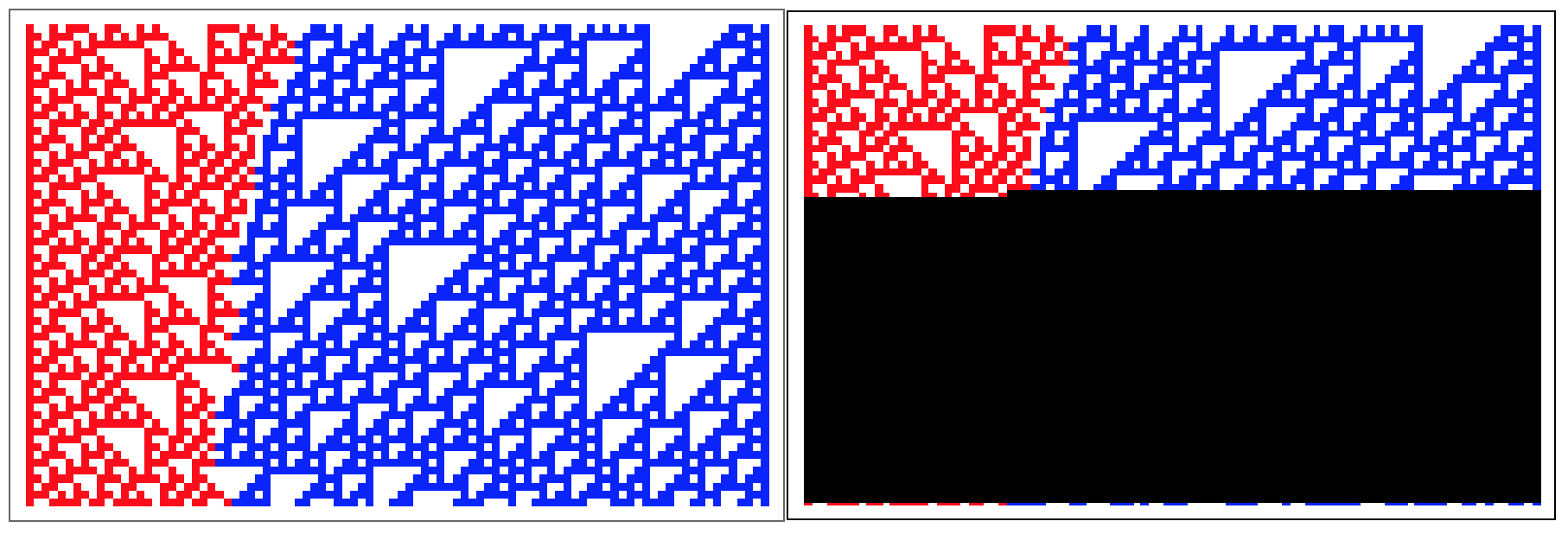}}\scalebox{.21}{\includegraphics{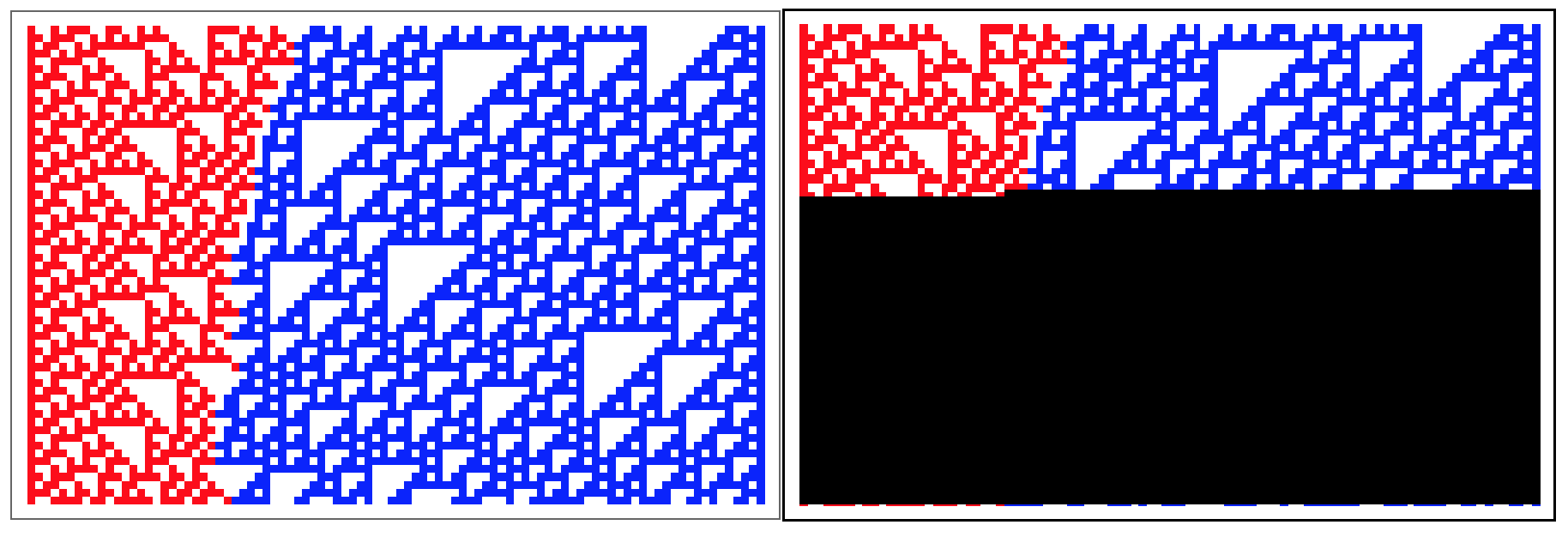}}

\caption{\label{si2b}A,B, and C are 3 additional non-selected random cases of decomposition by causal deconvolution (all involving rules 60 and 110 with non-trivial interaction with rule 531441) as introduced in this paper and with the same cutoff value. Each pair shows original (left) and decomposed (right) images on 3 different CA interacting cases. On the decomposed versions black covers shows the algorithm success at isolating one region from another with high accuracy. D and E are implementations of 2 other methods based on a single case (B) but based on Partial Information Decomposition using (classical) Mutual Information and also Normalized Compression Distance. In this case, cutoff values produced a horizontal straight line in a random place because the methods were unable to find any right-left separation stable enough to separate the image vertically.}
\end{figure}

For our purposes, and to avoid any bias due to white background, we start systems (ECAs) from initial conditions spanning the width of the evolution producing a squared or rectangular space-time diagram Fig.~\ref{si1}C rather than the typical pyramidal evolution that start from simplest initial condition (a black cell) such as examples in Figs.~\ref{si1}A,B. Fig.~\ref{si2b} illustrates the algorithm success at decomposing different 2-system interactions/compositions even with similar qualitative behaviour and statistics thus making it difficult for other statistical methods to perform the same task.

The following source code implements an algorithm determining the way in which local CA are dictated and govern their intersection. It can be implemented with this Wolfram Language (Mathematica) code:

\begin{verbatim}
R[x_]:=Thread[Rule[{{-1,1,0},{-1,0,1},{-1,1,1},{1,-1,1},{1,-1,0},
{1,1,-1},{1,0,-1}, {0,1,-1},{0,-1,1}, {1,-1,-1},{-1,1,-1},
{-1,-1,1}},Flatten[Take[Tuples[{-1,0,1},12],{x,x}]]]];
Code[n_]:=BitXor[n,BitShiftRight[n]];
RuleCode[n_]:=IntegerDigits[Code[n],2]
\end{verbatim}

where $RuleCode[]$ retrieves a rule governing the interaction between any 2 CAs.

Alternative code can be found at the Wolfram Demonstrations website at \url{http://demonstrations.wolfram.com/CompetingCellularAutomata/}~\cite{joostdem}.

\subsection{Graph Generation for Deconvolution Examples}

The graphs used throughout this paper were generated using the Wolfram Language on the Mathematica platform using the function RandomGraph[] with uniform distribution (UniformGraphDistribution[]) for Erd\H{o}s-R\'enyi graphs and a scale-free distribution (BarabasiAlbertGraphDistribution[]) for the scale-free networks constructed by starting from a cycle graph of size 3 and a vertex of $k$ edges added at each step according to the preferential attachment algorithm~\cite{barabasi} following a distribution proportional to the vertex degree. All experiments were replicated 20 times and results were aggregated and averaged.

\subsection{Other Methods and Measures Comparison}

\begin{figure}[ht]
\centering
\textbf{A}\\
\scalebox{.34}{\includegraphics{60110b.png}}\\
\medskip

\textbf{B}\hspace{5.8cm} \textbf{C}\\
\scalebox{.29}{\includegraphics{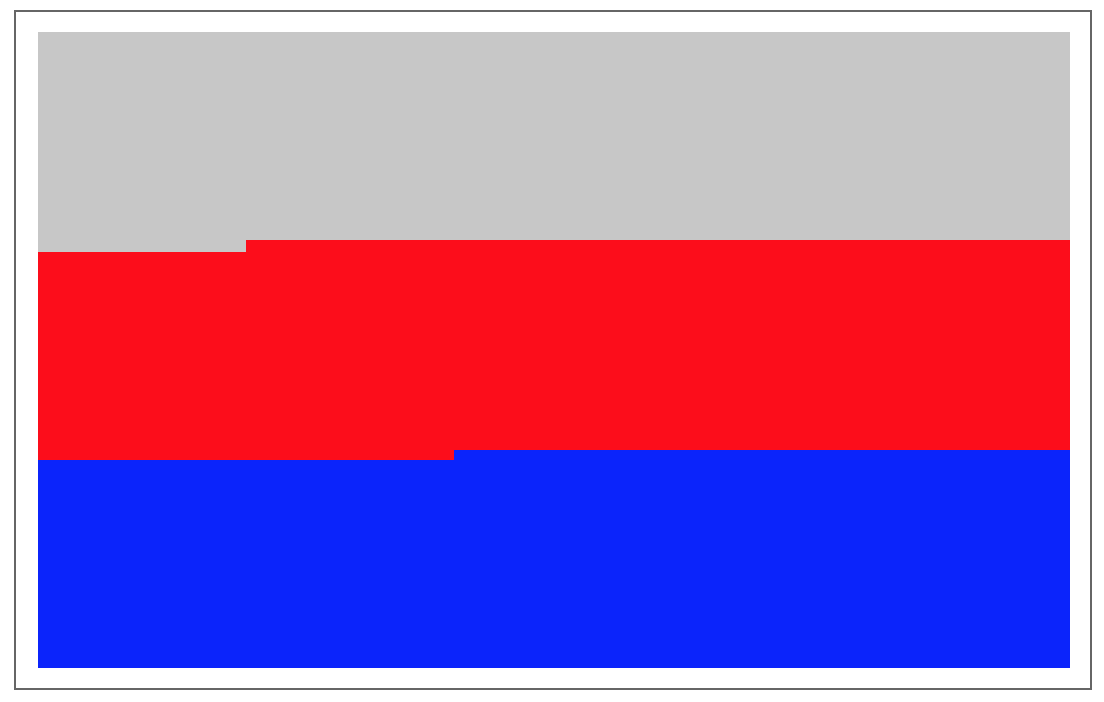}}\hspace{.6cm}\scalebox{.29}{\includegraphics{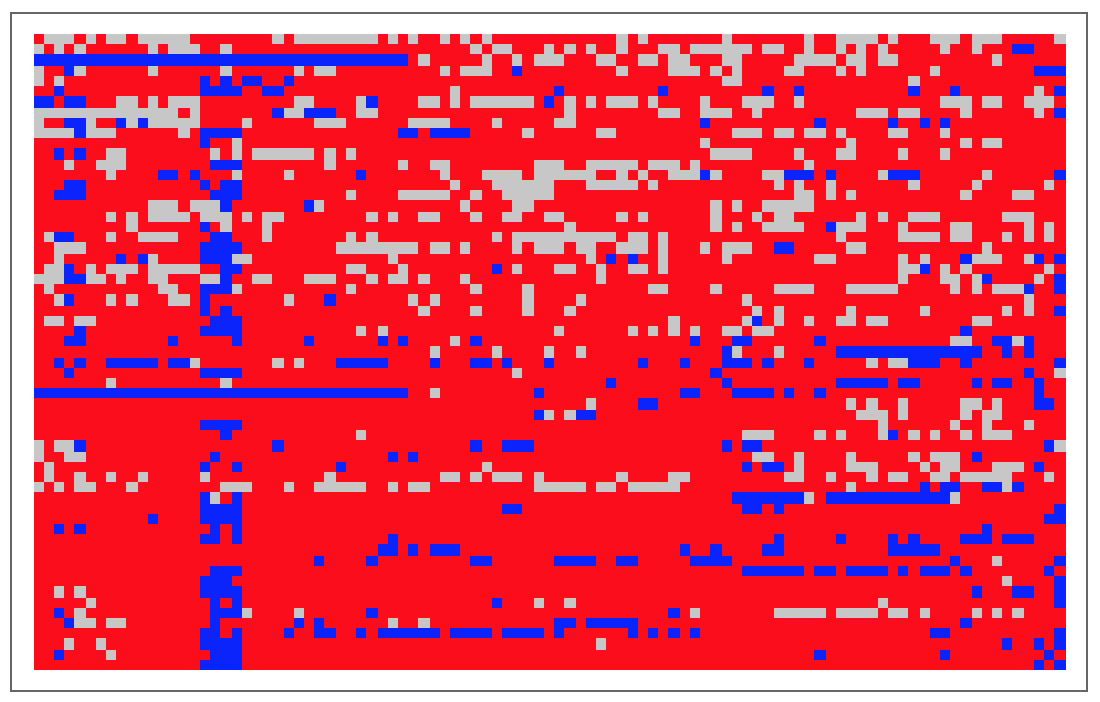}}

\caption{\label{si2}Original interaction (A), Shannon entropy (B) and lossless compression (Compress) (C) underperform, not being sensitive enough in performing the same task reported in Figs.~\ref{a}C-F. Interacting rule 531441.}
\end{figure}

Fig.~\ref{si2} shows the results obtained by replacing the algorithmic probability estimations (BDM) in our \textit{causal decomposition} method by classical information theory (Shannon entropy)~\ref{si2}B and one of the most popular lossless compression algorithms (Compress)~\ref{si2}C, based on LZW as an approximation to algorithmic (Kolmogorov-Chaitin) complexity instead of BDM. Because all values collapse into a single value for entropy, the colours displayed are the result of an artificial sorting of the pixels based on their indices, from top to bottom. Compression is a lower-quality approximation of what we reported in Figs.~\ref{a}C-F, where the reported algorithm based on the BDM is clearly an improvement.

\begin{figure}[ht!]
\centering
\textbf{A}\\
\scalebox{.38}{\includegraphics{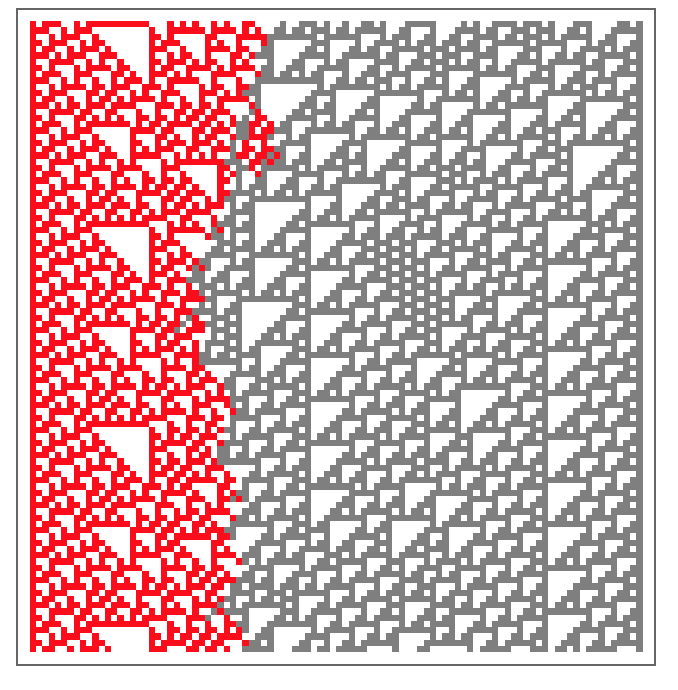}}\\
\medskip

\textbf{B}\hspace{3.8cm} \textbf{C}\hspace{3.8cm} \textbf{D}\\
\scalebox{.45}{\includegraphics{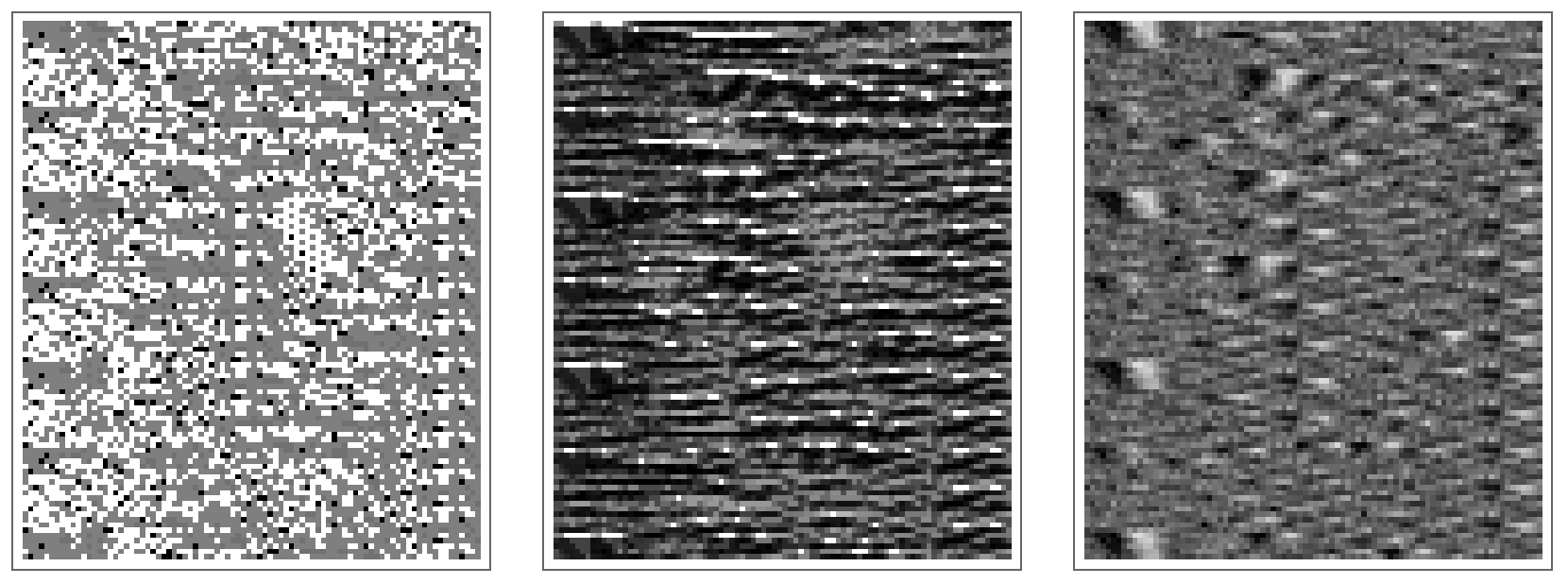}}

\caption{\label{si3}Sensitivity of methods applied to a 2-system interaction (A) both with similar qualitative properties based on (B) Partial Information Decomposition by Mutual Information (C) Normalised Compression Distance, and (D) algorithmic causal deconvolution as introduced in this paper. These arrays are similar to redundancy lattices indicating either synergy (whiter) or redundancy (darker).}
\end{figure}

In Fig.~\ref{si2b} and Fig.~\ref{si3}, we compared the accuracy and sensitivity of our \textit{causal deconvolution} algorithm introduced here against 2 other methods based on the concept of Partial Information Decomposition~\cite{williams} (PID) based on (classical) Mutual Information and also lossless compression by means of the Normalised Compression Distance~\cite{li2}~\cite{li2}.

The original PID~\cite{williams} method requires us to take all subsets of the \textit{predictor variables} in $X$, with $X$ set of all predictor variables of a target variable $S$ (e.g. the space-time diagram of interacting CA), and equivalent to our action of pixel perturbation.

Beyond the fact that the PID and NCD methods are based on different first principles, the main problem of both approaches seems to be a problem of sensitivity. In Fig.~\ref{si3}, we tested how many values (grey shades) each method was able to produce as an indication of how much each method would fine or coarse-grain the original 2-system interaction.

It was clear that the original version of PID was intractable, as it was impossible to deal with all the subsets of pixel perturbations of the interacting CA space-time diagram in Fig.~\ref{si3}A even for only 100 steps. So we proceeded by applying perturbations to rows only taking subsets of 6 bits at a time in a sliding non-overlapping window traversing each row left to right. 

The original PID algorithm is based on (classical) Mutual Information (thus Shannon entropy). $I(X; S)$ then  provides values for redundant and synergistic information between all subsets of $X$. The results in Fig.~\ref{si3} show the sensitivity of all 3 methods, demonstrating that all can capture features of the interactions, but with different sensitivity. The Normalised Compression Distance (Fig.~\ref{si3}C) as introduced in~\cite{li2}, and based on the algorithm Compress, in turn based on LZW, was the least sensitive, retrieving less than 5 different values to recolour the system thus heavily coarse-graining most features of Fig.~\ref{si3}A. Neither this adapted version of PID nor the version with Normalized Compression Distance where able to separate regions as done in~\ref{b}.

Causal deconvolution (Fig.~\ref{si3}D) was the most sensitive capturing more structure from the original 2-system interaction by virtue of finer grained capabilities. Moreover, as established in the main text, the underlying principles of each method are fundamentally different. Mutual Information, and thus Shannon entropy, can only quantify statistical regularities, while popular lossless compression is heavily related to Shannon entropy~\cite{emergence} despite its wide use as a method for approximating algorithmic complexity. The methods were not able to separate the interacting CA, as our method did (Fig.~\ref{b}), due to lack of sensitivity and specificity.

\subsection{Main Functions in Wolfram Language}

\begin{verbatim}
CausalDeconvolution[array_] := 
 Module[{pointrowmutation = 
    Flatten[Table[
      ReplacePart[array, {{i, j} -> Mod[array[[i, j]] + 1, 2]}], {i, 
       Length[array]}, {j, Length[array[[1]]]}], 1]}, 
  Reverse[SortBy[
    Thread[{Range[Length[pointrowmutation]], 
      BDM[array, 4] - (N /@ BDM[#, 4] & /@ pointrowmutation)}], 
    Last]]]
\end{verbatim}

\begin{verbatim}
PIDMI[array_] := 
 Module[{pointrowmutation = 
    Flatten[Table[
      ReplacePart[array, {{i, j} -> Mod[array[[i, j]] + 1, 2]}], {i, 
       Length[array]}, {j, Length[array[[1]]]}], 1]}, 
  Reverse[SortBy[
    Thread[{Range[Length[pointrowmutation]], 
      MutualInformation[#, array] & /@ pointrowmutation}], Last]]]
\end{verbatim}

\begin{verbatim}
PIDNCD[array_] := 
 Module[{pointrowmutation = 
    Flatten[Table[
      ReplacePart[array, {{i, j} -> Mod[array[[i, j]] + 1, 2]}], {i, 
       Length[array]}, {j, Length[array[[1]]]}], 1]}, 
  Reverse[SortBy[
    Thread[{Range[Length[pointrowmutation]], 
      NCD[#, array] & /@ pointrowmutation}], Last]]]
\end{verbatim}

\begin{verbatim}
MutualInformation[x_, y_] := 
 N[Entropy[x] + Entropy[y] - 
   Statistics`Library`NConditionalEntropy[x, y]]
\end{verbatim}

\begin{verbatim}
NCD[x_, y_] :=
N@Block[{
    cx = ByteCount[Compress[x]],
    cy = ByteCount[Compress[y]],
    cxy = ByteCount[Compress[Join[x, y]]]
    },
   (cxy - Min[cx, cy])/Max[cx, cy]
   ]
\end{verbatim}

\begin{verbatim}
CalculateInformationRowBDM[array_] := 
 Table[Table[
   Abs[StringBDM[
      StringJoin[ToString /@ Take[array[[i]], {m, m + 6}]]] - 
     StringBDM[
      StringJoin[ToString /@ Take[array[[i]], {m + 7, m + 12}]]]], {m,
     1, Length[array[[i]]] - 12}], {i, Length[array]}]
\end{verbatim}

\begin{verbatim}
CalculateInformationRowMI[array_] := 
 Table[Table[
   MutualInformation[Take[array[[i]], {m, m + 6}], 
    Take[array[[i]], {m + 7, m + 12}]], {m, 1, 
    Length[array[[i]]] - 12}], {i, Length[array]}]
\end{verbatim}

\begin{verbatim}
CalculateInformationRowNCD[array_] := 
 Table[Table[
   NCD[Take[array[[i]], {m, m + 6}], 
    Take[array[[i]], {m + 7, m + 12}]], {m, 1, 
    Length[array[[i]]] - 12}], {i, Length[array]}]
\end{verbatim}

\end{document}